\documentclass[pdflatex,sn-mathphys-num]{sn-jnl}

\usepackage{graphicx}%
\usepackage{multirow}%
\usepackage{amsmath,amssymb,amsfonts}%
\usepackage{amsthm}%
\usepackage{mathrsfs}%
\usepackage[title]{appendix}%
\usepackage{xcolor}%
\usepackage{textcomp}%
\usepackage{manyfoot}%
\usepackage{booktabs}%
\usepackage{algorithm}%
\usepackage{algorithmicx}%
\usepackage{algpseudocode}%
\usepackage{listings}%
\usepackage{placeins}%
\graphicspath{{figures/}}%
\usepackage{subcaption}
\usepackage{bbm}

\theoremstyle{thmstyleone}%
\theoremstyle{thmstyletwo}%

\theoremstyle{thmstylethree}%

\begin{document}

\title[Title]{Ensemble-Based Dirichlet Modeling for Predictive Uncertainty and Selective Classification}

\author*[1]{\fnm{Courtney} \sur{Franzen} \href{https://orcid.org/0009-0000-1543-2912}
{\includegraphics[width=0.2cm]{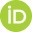}}}\email{courtney.franzen@ucdenver.edu}

\author[2]{\fnm{Farhad} \sur{Pourkamali-Anaraki}\href{https://orcid.org/0000-0003-4078-1676}
{\includegraphics[width=0.2cm]{ORCID-iD_icon_32x32.jpg}}}\email{farhad.pourkamali@ucdenver.edu}

\affil*[1]{\orgdiv{Department of Mathematical and Statistical Science}, \orgname{University of Colorado Denver}, \orgaddress{\city{Denver}, \postcode{80208}, \state{CO}, \country{USA}},}

\affil[2]{\orgdiv{Department of Mathematical and Statistical Science}, \orgname{University of Colorado Denver}, \orgaddress{\city{Denver}, \postcode{80208}, \state{CO}, \country{USA}}}


\abstract{
Neural network classifiers trained with cross-entropy loss achieve strong predictive accuracy but lack the capability to provide inherent predictive uncertainty estimates, thus requiring external techniques to obtain these estimates. In addition, softmax scores for the true class can vary substantially across independent training runs, which limits the reliability of uncertainty-based decisions in downstream tasks. Evidential Deep Learning aims to address these limitations by producing uncertainty estimates in a single pass, but evidential training is highly sensitive to design choices including loss formulation, prior regularization, and activation functions. Therefore, this work introduces an alternative Dirichlet parameter estimation strategy by applying a method of moments estimator to ensembles of softmax outputs, with an optional maximum-likelihood refinement step. This ensemble-based construction decouples uncertainty estimation from the fragile evidential loss design while also mitigating the variability of single-run cross-entropy training, producing explicit Dirichlet predictive distributions. Across multiple datasets, we show that the improved stability and predictive uncertainty behavior of these ensemble-derived Dirichlet estimates translate into stronger performance in downstream uncertainty-guided applications such as prediction confidence scoring and selective classification. 
}


\maketitle

\section{Introduction}\label{sec:momintro}

\begin{figure}[!t]
    \centering
    \includegraphics[width=0.8\textwidth]{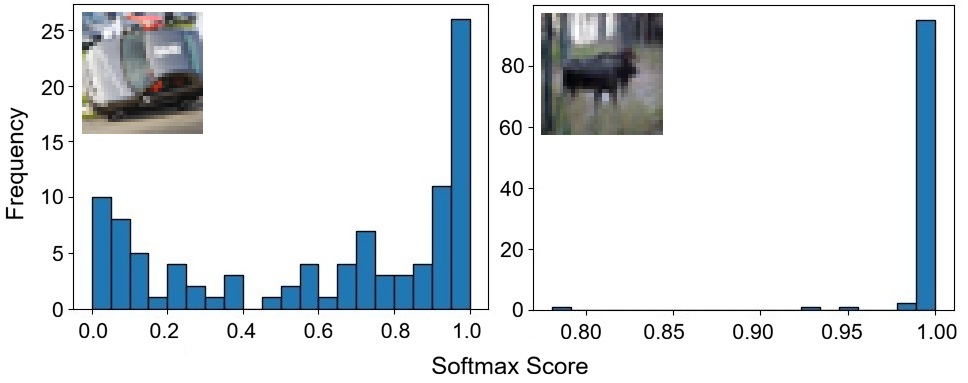}
    \caption[Frequency Plots of Softmax Scores for the True Class]{
    Frequency plots of softmax scores for the true class for two representative samples from a 10-class image classification task across 100 randomly initialized classifiers trained with a cross-entropy loss function. The left plot (automobile) shows a difficult sample with highly variable scores, while the right plot (deer) shows an easier sample whose scores remain concentrated but still exhibit nontrivial variability.
    }
    \label{fig:softmaxfreq}
\end{figure}

Deploying classifiers in high-stakes domains requires principled estimates of predictive uncertainty to support reliable decision making. In standard classification systems trained with cross-entropy (CE) loss, the output consists of softmax scores rather than explicit uncertainty estimates. Although these scores are often interpreted as measures of predictive confidence, they are not guaranteed to reflect calibrated correctness and do not provide explicit measures of uncertainty~\cite{guo2017calibration, tomani2021fast, molchanov2017variational, kingma2015variational, gal2016dropout}. Moreover, the magnitude of softmax scores can vary substantially across independent training runs, even when predicted class labels remain unchanged~\cite{lakshminarayanan2017simple, ashukha2021pitfalls, monarch2021human}. Fig.~\ref{fig:softmaxfreq} illustrates this phenomenon on two representative samples from a 10-class image dataset, where identical inputs yield noticeably different probability estimates under different random initializations.  Taken together, these limitations highlight the need for predictive representations that encode uncertainty more directly and consistently than single-run softmax outputs.

Evidential Deep Learning (EDL) aims to address this need by modeling class probabilities as a Dirichlet-distributed random variable~\citep{sensoy2018evidential, ulmer2023posterior}. Grounded in Subjective Logic~\citep{josang2016subjective}, the approach interprets the Dirichlet distribution as representing uncertainty over categorical probabilities relative to a fixed reference distribution (the \emph{base rate}), which encodes the prior probability of each class in the absence of evidence.  Rather than introducing parameter stochasticity to capture predictive uncertainty, an EDL network deterministically outputs nonnegative evidence values that define Dirichlet concentration parameters for each input. These concentration parameters specify a predictive distribution over categorical likelihoods, yielding both a predictive mean and an associated variance in a single pass. In this formulation, uncertainty is encoded directly in the predictive distribution itself.

Despite its theoretical appeal, EDL suffers from practical challenges. Its performance is sensitive to design choices such as loss function, prior regularization, and evidence activation~\cite{sensoy2021risk, jurgens2024faithful, davies2023EDLsignals, shen2024mirage, malinin2018prior}. These choices can dramatically affect the Dirichlet concentration parameters and, consequently, the resulting predictive variance. Some configurations lead to unstable training, collapsed uncertainty, or unreliable outputs, particularly in fine-grained or high-class-count tasks. These issues limit the robustness and generalizability of EDL across datasets~\cite{pandey2025gen}, motivating the need for fallback or complementary methods when evidential training fails.

\begin{figure}[!t]
    \centering
    \includegraphics[width=\textwidth]{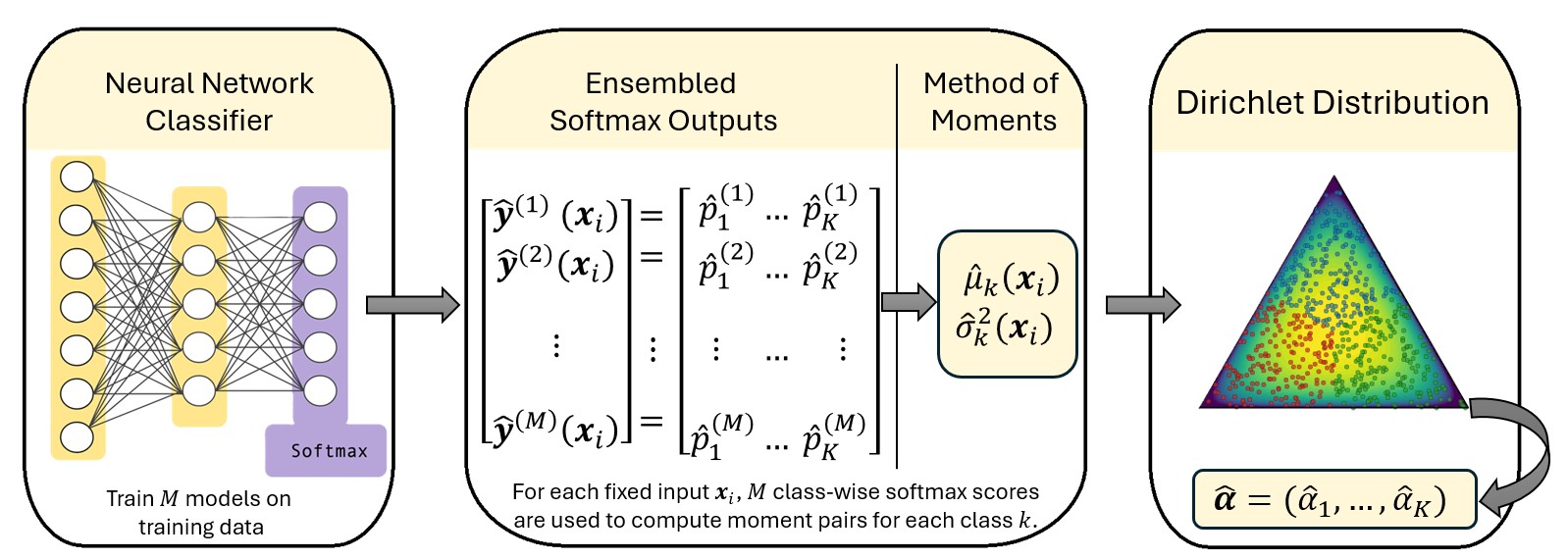}
    \caption[Overview of the Method of Moments Dirichlet Construction]{Overview of the method of moments Dirichlet construction. An ensemble of CE-trained models produces multiple softmax probability vectors for a fixed input. These are treated as empirical samples on the probability simplex, from which class-wise means and variances are computed and matched to a Dirichlet distribution, yielding per-input concentration parameters $\hat{\boldsymbol{\alpha}}$.}
    \label{fig:ensemble_flow_figure}
\end{figure}

To address this, we propose a simple and effective alternative for generating Dirichlet-based uncertainty estimates: a method of moments (MoM) construction with optional maximum likelihood refinement that fits a Dirichlet distribution to the empirical distribution of softmax outputs from an ensemble of CE-trained models. Fig.~\ref{fig:ensemble_flow_figure} shows how repeated softmax vectors for a fixed input can be treated as samples on the probability simplex. By matching class-wise empirical means and variances to a Dirichlet distribution, we obtain per-input concentration parameters that encode input-dependent uncertainty without requiring evidential training.

This MoM-derived Dirichlet construction produces predictive distributions that match EDL in interpretability and application, enabling direct comparison in downstream tasks such as calibration, confidence ranking, and selective classification. In particular, we show that total Dirichlet variance can serve as an informative ordering statistic for abstention, and that MoM-derived models can achieve superior performance on variance-based selective classification metrics compared to EDL configurations.

In total, this work makes the following contributions:
\begin{enumerate}
    \item We introduce a method of moments estimator for constructing Dirichlet distributions from ensembles of CE–trained models, yielding input-dependent concentration parameters without specialized evidential loss design or unstable training dynamics.
    \item We investigate an optional maximum likelihood refinement via fixed-point iteration, evaluating whether likelihood-based optimization meaningfully improves moment-based Dirichlet estimates in calibration and selective classification settings.
    \item We provide a systematic empirical analysis of Dirichlet-based classifiers for uncertainty-aware decision making, showing that evidential design choices induce substantial variation in confidence behavior, calibration diagnostics, and selective performance across otherwise similar models.
    \item We empirically demonstrate that ensemble-derived Dirichlet variance serves as an effective ordering statistic for selective classification, achieving competitive or superior retained accuracy, F1 score, and negative log-likelihood under matched risk constraints
\end{enumerate}

The remainder of the paper is organized as follows. Section~\ref{sec:mombackground} reviews Dirichlet-based uncertainty modeling and confidence calibration analysis. Section~\ref{sec:mommethodology} presents the ensemble-based Dirichlet construction and its optional maximum likelihood refinement, followed by the variance-based selective classification framework. Section~\ref{sec:momresults} reports three experimental studies: calibration analysis across evidential formulations, comparison of moment-based and likelihood-refined Dirichlet estimators against evidential baselines, and a variance-based selective classification study evaluating risk–coverage performance under fixed risk levels. Finally, Section~\ref{sec:momdiscussion} discusses implications for uncertainty modeling and outlines directions for future work.

\section{Background}
\label{sec:mombackground}

\subsection{Evidential Deep Learning}\label{sec:edl}

In classification tasks, neural networks typically produce classification scores using a final linear layer defined as 
\begin{equation}\label{eq:logits}
    \mathbf{z} := \mathbf{W} \mathbf{h} + \mathbf{b} \in \mathbb{R}^K,
\end{equation}
where $\mathbf{h} \in \mathbb{R}^d$ is the final hidden layer feature vector, $\mathbf{W} \in \mathbb{R}^{K \times d}$ is the output weight matrix, and $\mathbf{b} \in \mathbb{R}^K$ is the bias vector. These logits $\mathbf{z}$ are then transformed (via a model-specific mapping) into a vector of prediction scores $\hat{\mathbf{p}} = [\hat{p}_1, \dots, \hat{p}_K]$ over the $K$ classes, where each $\hat{p}_k$ is often interpreted as the probability that the sample belongs to class $k$. Each ground-truth label is encoded as a \textit{one-hot} vector $\mathbf{y} \in \{0,1\}^K$ with a single 1 at the true class index $j$ (i.e., $y_j = 1$ and $y_k = 0$ for all $k \ne j$), such that $\sum_{k=1}^K y_k = 1$.

The standard approach for training classifiers is to minimize the cross-entropy loss function. For a single training sample with one-hot label vector $\mathbf{y}$ and predicted class probabilities $\hat{\mathbf{p}}$, the cross-entropy loss is

\begin{equation}\label{eq:celoss}
    \mathcal{L}_{\mathrm{CE}} = -\sum_{k=1}^{K} y_k \log \hat{p}_k ,
\end{equation}
and the full loss function is the average of this loss over the training dataset.  Class prediction scores are obtained by applying the standard softmax mapping to the logits, producing a vector of prediction scores

\begin{equation}\label{eq:softmax}
    \hat{p}_k = \frac{\exp(z_k / T)}{\sum_{j=1}^{K} \exp(z_j / T)}.
\end{equation}

During training, this mapping is used without applying the temperature scaling parameter $T$ to the logits.  During calibration, the same mapping may be optionally reused with $T>0$ tuned post hoc on a held-out calibration dataset to improve probability calibration without affecting classification accuracy~\cite{monarch2021human}.

Because these scores form a valid probability distribution over classes (i.e., each $\hat{p}_k \in [0,1]$ and $\sum_k \hat{p}_k = 1$), softmax outputs are often interpreted as measures of predictive confidence. While this formulation is effective for optimizing classification accuracy, the resulting scores are not guaranteed to be calibrated estimates of predictive correctness (i.e., predicted confidence values do not necessarily match the true empirical frequency of correct predictions) and can exhibit substantial variability across repeated training runs, as shown in Fig.~\ref{fig:softmaxfreq}. This instability stems from sensitivity to random initialization, data order, and optimization noise, leading to inconsistent predictive behavior even for the same input~\cite{lakshminarayanan2017simple, ashukha2021pitfalls, monarch2021human}. As a result, naive use of softmax scores as confidence can obscure a model’s true reliability and understanding of uncertainty~\cite{guo2017calibration, tomani2021fast, wang2023calibrated}. To mitigate these effects, practitioners commonly apply post hoc calibration methods such as temperature scaling in \eqref{eq:softmax}. However, such approaches introduce additional complexity, require a held-out calibration set, and are known to degrade under distribution shift~\cite{ovadia2019trust}, limiting their usefulness in practical deployments.

In contrast, Evidential Deep Learning draws on Subjective Logic~\cite{josang2016subjective, dst2007classic} to reinterpret classification as the quantification of opinions over categorical outcomes, enabling a principled connection between neural network outputs and Dirichlet distributions. In this framework, Dirichlet-based classifiers output \textit{concentration parameters} $\boldsymbol{\alpha} = (\alpha_1, \dots, \alpha_K)$ of a Dirichlet distribution $\mathrm{Dir}(\boldsymbol{\alpha})$ over class probabilities. With the initial development of EDL, each concentration parameter was defined as $\alpha_k := e_k + 1$, where $e_k$ denotes the subjective \textit{evidence mass} inferred by the network for class $k$.  This evidence is obtained by applying a nonnegative transformation (e.g., ReLU, softplus, or exponential) to the network output $z_k$. The \emph{total evidence} $\alpha_0$ captures the overall strength of belief, and its relative magnitude governs the certainty of the resulting prediction~\cite{josang2016subjective}. As a result, the model encodes both the predictive mean and its associated uncertainty within a single Dirichlet distribution over class probabilities $\boldsymbol{p} = [p_1, \ldots, p_K]$ on the $(K{-}1)$-simplex, whose density is given by

\begin{equation}\label{eq:dirpdf}
    \mathrm{Dir}(\boldsymbol{p} \mid \boldsymbol{\alpha}) 
    = \frac{1}{\mathrm{B}(\boldsymbol{\alpha})} \prod_{k=1}^K p_k^{\alpha_k - 1}, 
\end{equation}

\begin{equation}\label{eq:mvbfunc}
    \text{where}  \; \mathrm{B}(\boldsymbol{\alpha}) := \frac{\prod_{k=1}^K \Gamma(\alpha_k)}{\Gamma(\alpha_0)} ,\; \alpha_0 := \sum_k \alpha_k
\end{equation}

While EDL does not perform full Bayesian inference, it borrows both structure and intuition from classical Bayesian methods. In particular, the Dirichlet distribution arises as the conjugate prior to the Categorical likelihood in multi-class classification problems, providing a principled representation of uncertainty over discrete outcomes.  Consequently, unlike the single point estimates produced by CE-based classifiers, a Dirichlet distribution represents a distribution \emph{over} categorical probability vectors.  This distinction is critical: the predictive object of interest is no longer a heuristic score, but a random variable whose moments have a direct probabilistic interpretation.  In particular, EDL models produce a predictive mean and variance for class $k$ given by 
\begin{equation}\label{eq:meanvar}
    \mathbb{E}[p_k \mid \boldsymbol{\alpha}] = \frac{\alpha_k}{\alpha_0}, \quad \mathbb{V}[p_k \mid \boldsymbol{\alpha}] = \frac{\alpha_k\left(\alpha_0-\alpha_k\right)}{\alpha_0^2\left(\alpha_0+1\right)}. 
\end{equation}

Building on these foundations, a range of formulations have emerged within the EDL framework. We examine several representative configurations here.  \citet{sensoy2018evidential} proposed training Dirichlet-based classifiers by minimizing the expected value of a standard loss function under the predictive Dirichlet distribution. One of the three loss functions introduced in their work, the mean squared error (MSE) formulation directly targets the predictive mean and the one-hot target. Evaluating this expectation under the Dirichlet distribution yields a bias–variance decomposition of the per-sample loss:
\begin{align}\label{eq:mseloss}
    \mathcal{L}_{\mathrm{MSE}}
    &= \mathbb{E}_{\mathbf{p} \sim \mathrm{Dir}(\boldsymbol{\alpha})} \left[ \| \mathbf{y} - \mathbf{p} \|^2_2 \right] \notag \\
    &= \sum_{k=1}^K \left[\left( y_k - \frac{\alpha_k}{\alpha_0} \right)^2 + \frac{\alpha_k(\alpha_0 - \alpha_k)}{\alpha_0^2(\alpha_0+1)}\right].
\end{align}

A second variant adopts a more conventional classification objective for comparison. Specifically, the expectation under the predictive Dirichlet distribution is applied to the standard cross-entropy loss function from  \eqref{eq:celoss}, yielding the so-called digamma per-sample loss,
\begin{align}\label{eq:digammaloss}
    \mathcal{L}_{\mathrm{Digamma}} 
    &= \mathbb{E}_{\mathbf{p} \sim \mathrm{Dir}(\boldsymbol{\alpha})}
    \left[-\sum_{k=1}^K y_k \log p_k\right] \notag \\
    &= \sum_{k=1}^K y_k \left(\psi(\alpha_0) - \psi(\alpha_k)\right),
\end{align}
where $\psi(\cdot)$ denotes the digamma function. The digamma function is defined as the derivative of the logarithm of the gamma function $\Gamma(\cdot)$,
\begin{equation}\label{eq:digammafunc}
    \psi(u) := \frac{d}{du}\log \Gamma(u),
\end{equation}
For completeness, full derivations of the simplified forms of \eqref{eq:mseloss} and~\eqref{eq:digammaloss} are provided in Appendix~\ref{sec:lossders}.

To mitigate overconfident predictions and uncontrolled growth in the total concentration parameter, regularization terms are introduced. A common approach adds a Kullback–Leibler (KL) divergence between the predicted Dirichlet distribution and a uniform prior $\mathrm{Dir}(\mathbf{1})$, where $\mathbf{1}$ denotes the vector of all ones. Applying this regularization term to the loss in \eqref{eq:mseloss}, we get the final per-sample loss of 
\begin{align}\label{eq:dirloss}
    \mathcal{L}_{\mathrm{MSE+KL}} = \sum_{k=1}^K \left( (y_k - \hat{p}_k)^2 + \frac{\hat{p}_k(1 - \hat{p}_k)}{\alpha_0 + 1} \right) \notag \\
    + \lambda_{\mathrm{KL}} \cdot \mathrm{KL}\left[ \mathrm{Dir}(\boldsymbol{\alpha}) \; \| \;  \mathrm{Dir}(\mathbf{1}) \right],
\end{align}
which we will refer to as the ``MSE+KL'' loss.

In the context of EDL, the uniform prior $\mathrm{Dir}(\mathbf{1})$ establishes a neutral, uncertainty-preserving baseline and enables a simplified closed-form expression for the KL divergence:
\begin{align}
    \mathrm{KL}\left[ \mathrm{Dir}(\boldsymbol{\alpha}) \; \| \; \mathrm{Dir}(\mathbf{1}) \right] =
    \log \left( \frac{ \Gamma\left( \alpha_0 \right) }
    { \Gamma(K) \prod_{k=1}^K \Gamma(\alpha_k) } \right)
    \notag \\ + \sum_{k=1}^K (\alpha_k - 1)\left( \psi(\alpha_k) - \psi(\alpha_0) \right).
\end{align}

Because this KL penalty can dominate learning in early training, \citet{sensoy2024risk} propose a warm-up schedule $\lambda_{\mathrm{KL}} := \min(1, t/T)$, where $t$ denotes the current epoch and $T$ is a warm-up threshold (typically 10 epochs). In contrast, \citet{franzen2025dirclassifiers} introduce a fully linear annealing schedule inspired by~\cite{fu2019cyclical, sonderby2016ladder, kingma2015variational, he2019lagging} that is normalized by the number of classes,
\begin{equation}\label{eq:annealkl}
    \lambda_{\mathrm{KL}} = \frac{\lambda_0}{K}\cdot\frac{t}{T},
\end{equation}
where $t$ denotes the current epoch, $T$ represents the total number of training epochs, and $\lambda_0$ controls the maximum effective strength of the KL regularizer at the end of training.  This fully linear annealing schedule mitigates early training instability and uncontrolled evidence growth in high-class-count settings and empirically promotes stable training behavior for the MSE+KL loss formulation \cite{franzen2025dirclassifiers}.

In addition to KL-based regularization, another common regularization approach introduces explicit penalties on the magnitude of total accumulated evidence $\alpha_0$ \cite{ulmer2023posterior}. A commonly used example is the log-evidence regularizer,
\begin{equation}
\mathcal{L}_{\mathrm{LOGEV}} = \lambda_{\mathrm{EV}} \log \left(1 + \alpha_0\right).
\end{equation}
Unlike the KL divergence in \eqref{eq:mseloss}, this term operates directly on the magnitude of accumulated evidence rather than enforcing agreement with a reference Dirichlet distribution.

\begin{figure}[t]
    \centering
    \includegraphics[width=0.6\linewidth]{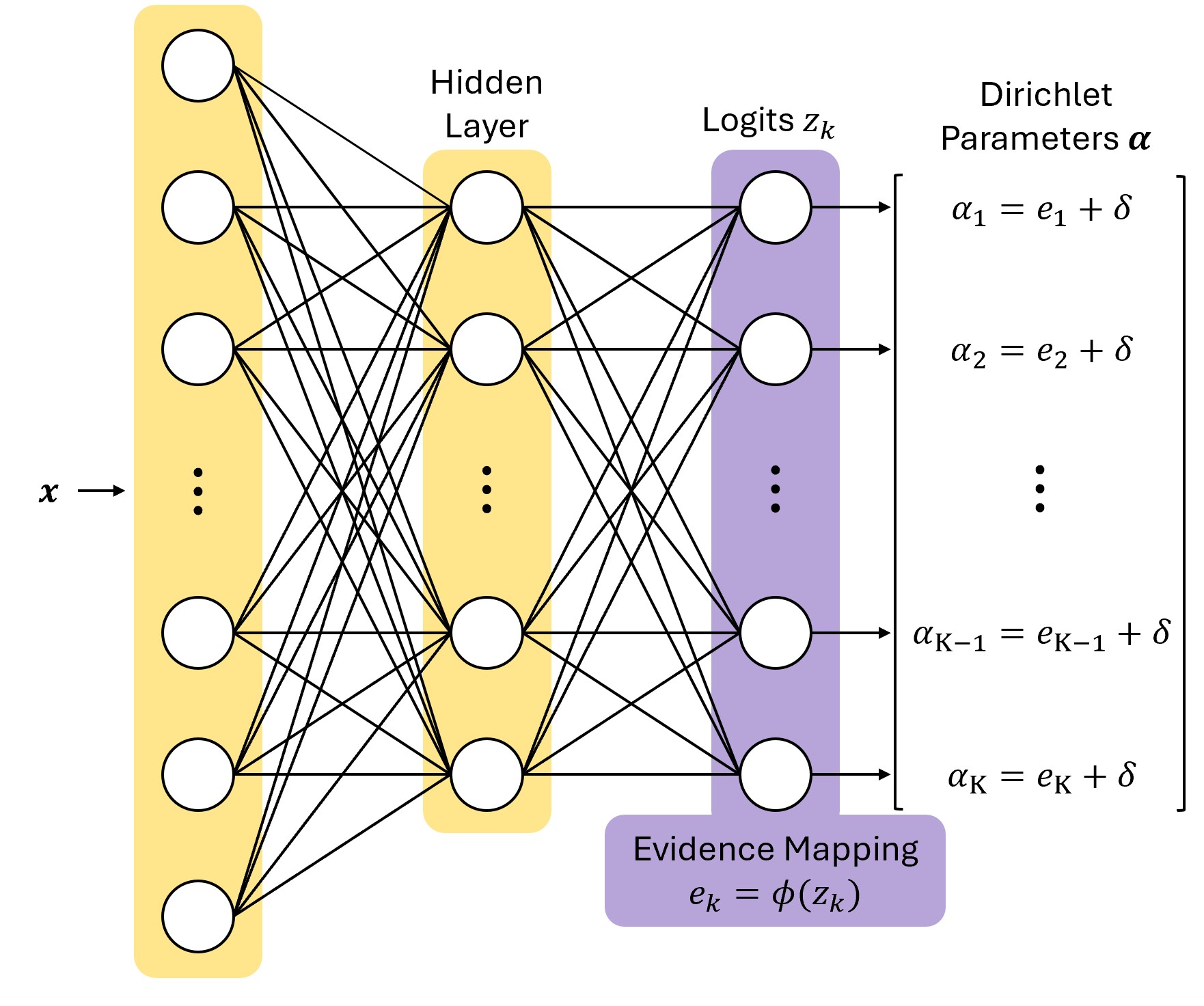}
    \caption[Diagram of an EDL Network.]{Diagram of an EDL Network. A neural network maps an input feature vector $\mathbf{x}$ to class-specific logits, which are transformed by a nonnegative activation function to produce evidence values. These evidence values are converted into Dirichlet concentration parameters, optionally including an additive prior offset. The resulting parameter vector defines a Dirichlet distribution over categorical class probabilities from which predictive confidence and uncertainty measures can be derived.
    }
\label{fig:edl_arch}
\end{figure}

Finally, we consider three parameterizations $\phi(\cdot)$ for mapping logits to nonnegative evidence used to construct Dirichlet concentration parameters. As illustrated in Fig.~\ref{fig:edl_arch}, a neural network maps an input vector $\mathbf{x}$ to class-specific logits $z_k$. A nonnegative activation function then converts these logits to evidence values $e_k = \phi(z_k)$, and the final class-wise concentration is obtained from $\alpha_k = e_k + \delta$, where $\delta \in \{0,1\}$ denotes a binary offset controlling the inclusion of an additive unit prior. Setting $\delta = 1$ recovers the standard EDL formulation~\cite{sensoy2024risk}, while $\delta = 0$ yields an unshifted parameterization.
The activation functions evaluated in this work are:
\begin{itemize}
    \item \textbf{SoftPlus}~\cite{ulmer2023posterior}, a smooth, nonnegative nonlinearity,
        \begin{equation}
        \phi_{\mathrm{SP}}(z_k) = \log \left(1 + \exp(z_k)\right).
        \end{equation}
        
    \item \textbf{Adaptive Softplus}, an adaptive softplus variant with learnable scale and growth parameters $\beta_k \geq 1$ and $\gamma_k > 0$ inspired by the work in \cite{pourkamali2024adaptive},
        \begin{equation}
        \phi_{\mathrm{SA}}(z_k) = \log \left(\beta_k + \gamma_k \exp(z_k)\right),
    \end{equation}
    where $\beta_k$ and $\gamma_k$ are optimized jointly with the network weights, allowing the model to regulate evidence growth dynamically during training.
        
    \item \textbf{Exponential}, corresponding to the evidential formulation adopted in recent EDL work~\cite{sensoy2024risk},
    \begin{equation}
        \phi_{\mathrm{EXP}}(z_k) = \exp(z_k),
    \end{equation}
    which can result in aggressive evidence growth and sensitivity to outliers. To mitigate numerical overflow, logits are clamped prior to exponentiation with gradients preserved via the original pre-clamped inputs.
\end{itemize}

\subsection{Confidence Calibration Analysis}\label{sec:momcalibration}

Deep learning classifiers are increasingly deployed in high-stakes domains such as medical diagnosis, autonomous systems, and safety-critical decision making~\cite{gilany2022prostate, hendrix2024fedev, khot2024stormEDL, li2022brainEDL, singh2021evidential_graph, yu2024bevuncertainty, pourkamali2023evaluation}. In such settings, misclassifications are inevitable, but \emph{overconfident} errors (i.e., predictions that are both wrong and made with high certainty) are especially dangerous~\cite{sensoy2024risk,tomani2021fast}. Reliable confidence estimates are therefore essential, particularly when used to inform selective classification systems that defer uncertain predictions to human review or auxiliary models.

Following prior work~\cite{charpentier2020posterior, sensoy2018evidential, franzen2025dirclassifiers}, let $\{\mathbf{x}_i\}_{i=1}^{N}$ denote a collection of $N$ test input feature vectors with $\mathbf{x}_i \in \mathbb{R}^{p}$; these inputs may represent images, but we assume one-dimensional vectors for simplicity in the following discussion. We extract a scalar confidence score for each input $\mathbf{x}_i$, defined as the maximum predicted class probability,
\begin{equation}
    \hat{c}(\mathbf{x}_i) := \max_k \hat{p}_k(\mathbf{x}_i).
\end{equation}

To evaluate the reliability of these scores, EDL research commonly employs expected calibration error (ECE)~\cite{Naeini2015bayesianbinning}. ECE measures the discrepancy between predicted confidence and empirical accuracy by partitioning predictions into a fixed number of confidence bins. Within each bin, the average predicted confidence is compared to the fraction of correct predictions, and the absolute difference between these quantities is computed. The final ECE score is obtained as the weighted average of these differences across bins, with weights proportional to the number of samples assigned to each bin. However, relying on ECE as the sole metric for calibration analysis is problematic. ECE is a biased, bin-sensitive metric that assesses calibration only with respect to the maximum predicted probability and cannot be optimized directly~\cite{vaicenavicius2019calibration, kumar2020verified, nixon2020measuring, ashukha2021pitfalls}. These limitations are particularly severe in regimes where predictions become uncertain or degenerate.

\begin{figure}[!t]
    \centering
    \begin{subfigure}{\linewidth}
        \includegraphics[width=\textwidth]{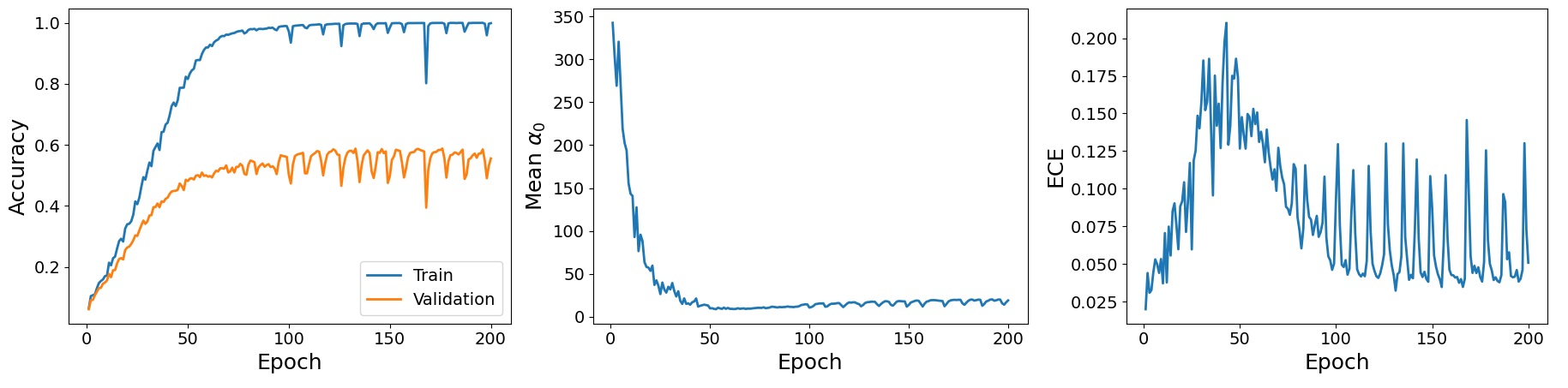}
        \caption{Successful training (with KL annealing)}
        \label{fig:success_plots}
    \end{subfigure}
    
    \begin{subfigure}{\linewidth}
        \includegraphics[width=\textwidth]{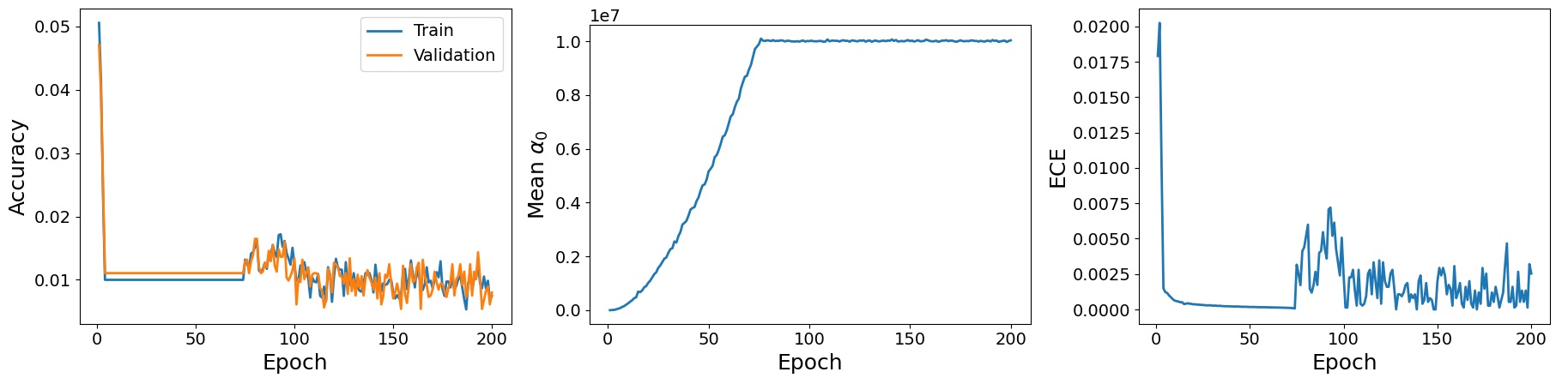}
        \caption{Training collapse (without annealing)}
        \label{fig:fail_plots}
    \end{subfigure}
    
    \caption[Successful Versus Collapsed Training Dynamics for Dirichlet Classifier]{Training dynamics for Dirichlet classifiers on CIFAR-100 trained from random initialization. With a linearly annealed KL term, MSE+KL training converges to stable accuracy with controlled growth of the total concentration parameter $\alpha_0$. Removing KL annealing induces collapse: accuracy degrades to chance ($1\%$ for $K=100$) while $\alpha_0$ grows explosively (mean $\alpha_0 \approx 10^{7}$), yielding degenerate confidence estimates despite decreasing loss. Notably, the collapsed regime achieves lower ECE values than the successful run ($\text{ECE}_{\text{fail}}$ $\approx$ $0.0025$ versus $\text{ECE}_{\text{succ}}$ $\approx$ $0.075$), which would typically suggest better calibration, illustrating that low ECE can mask training collapse.}
    \label{fig:fail_demo}
\end{figure}

Fig.~\ref{fig:fail_demo} illustrates a critical failure mode of Dirichlet classifiers in small-data, high-class-count settings. Both models share the same architecture, loss, and evidence parameterization; the only difference is whether the KL regularization term is sufficiently annealed during training. With annealing, accuracy improves steadily and the total concentration parameter $\alpha_0$ remains bounded, indicating controlled evidence accumulation. Without annealing, optimization exploits the loss by uniformly inflating evidence across classes: predictive means converge to the uniform distribution, accuracy collapses to chance, and $\alpha_0$ grows arbitarily large.

In this collapsed regime, ECE becomes actively misleading. As confidence scores are uniformly suppressed, ECE appears artificially small despite the model failing to produce either discriminative predictions or meaningful uncertainty estimates. The resulting low ECE reflects trivial underconfidence induced by a degenerate predictive distribution rather than reliable confidence behavior, underscoring the need for analyses that move beyond scalar calibration metrics.

Despite this failure mode, many EDL studies continue to report ECE as the primary or sole calibration metric, often without accompanying diagnostic analysis or visualization~\cite{tan2025evidentialpinn, yang2025bilevel, kandemir2022evidential, caprio2025credal, deery2023propan, grefsrud2025calibrated, agbelese2025megan, kim2026E2BKI, oh2021improving, bao2022opental}.  In contrast, only a small number of works explicitly examine calibration behavior using more informative diagnostics, such as reliability diagrams or analyses over the full predictive distribution~\cite{tan2025inferring, radev2021amortized, pandey2025gen}. These approaches emphasize the structure of confidence errors across the probability range rather than collapsing calibration behavior into a single scalar value.

These observations motivate a shift in how calibration is evaluated for Dirichlet classifiers. Rather than relying on scalar summary metrics such as ECE, we adopt reliability diagrams~\cite{degroot1983forecast, niculescu2005goodprobs} and confidence histograms~\cite{guo2017calibration} as primary diagnostic tools. These diagnostics directly characterize the relationship between predicted confidence and empirical correctness, allowing us to distinguish meaningful uncertainty estimation from degenerate underconfidence.

\begin{figure}
    \centering
    \includegraphics[width=\textwidth]{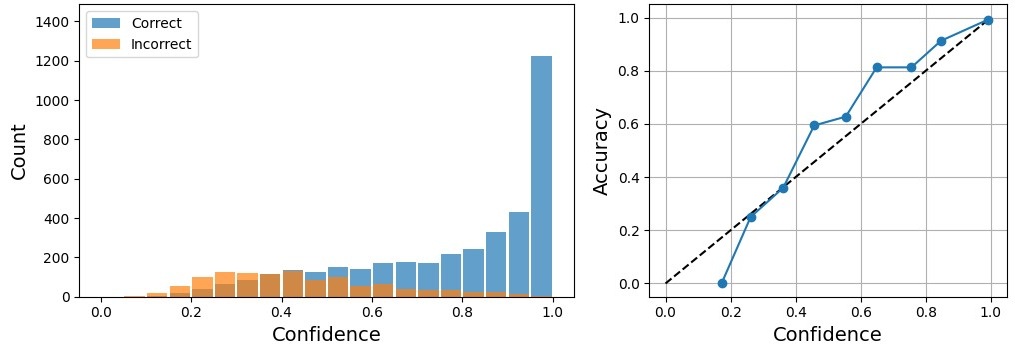}
    \caption[Example of Calibration Diagnostics]{Example confidence diagnostics. The confidence histogram (left) shows the distribution of confidence scores for correct and incorrect predictions. The reliability diagram (right) compares empirical accuracy to predicted confidence; the dashed diagonal denotes perfect calibration, with points above indicating underconfidence and below indicating overconfidence.}
    \label{fig:rel_hist_diagram}
\end{figure}

To reason formally about calibration, we begin with a probabilistic definition. Let $\hat{k}(\mathbf{x}_i)$ denote the predicted class index for input $\mathbf{x}_i$, and let $k(\mathbf{x}_i)$ denote the ground-truth class index. We define the prediction correctness random variable $R(\mathbf{x}_i)$ as
\begin{equation}
    R(\mathbf{x}_i) := \mathbbm{1}\{\hat{k}(\mathbf{x}_i) = k(\mathbf{x}_i)\},
\end{equation}
where $\mathbbm{1}\{\cdot\}$ denotes the indicator function, which takes value $1$ when the condition inside the braces is true and $0$ otherwise.
Let $\hat{c}(\mathbf{x}_i)\in[0,1]$ denote the scalar confidence score associated with the prediction at input $\mathbf{x}_i$. In line with \citet{degroot1983forecast}, the confidence score is said to be \emph{calibrated} if
\begin{equation}\label{eq:calibrated}
    \mathbb{P}(R = 1 \mid \hat{c} = c) = c \quad \text{for all } c \in [0,1].
\end{equation}
Under this definition, confidence enables a direct frequency interpretation: among all predictions assigned confidence $c$, the empirical fraction of correct predictions should be approximately $c$. To analyze this definition empirically, we employ reliability diagrams, which estimate the conditional probability $\mathbb{P}(R=1 \mid \hat{c}=c)$ from finite samples. Predictions are grouped into $B$ disjoint confidence bins
\begin{equation}
    I_b := \left( \frac{b-1}{B}, \frac{b}{B} \right], \qquad b = 1,\dots,B,
\end{equation}
where $B$ controls the resolution of the estimate; while reliability diagrams are sensitive to the choice of $B$, this affects only the granularity of the visualization rather than collapsing into a single scalar summary like with ECE.  The index set of samples whose confidence falls into bin $I_b$ is denoted
\begin{equation}
    \mathcal{S}_b := \{\; i : \hat{c}(\mathbf{x}_i) \in I_b \}.
\end{equation}
The empirical accuracy and average confidence within each bin are computed as
within each bin are computed as
\begin{align}
    \mathrm{acc}(\mathcal{S}_b) &:= \frac{1}{|\mathcal{S}_b|} 
    \sum_{i \in \mathcal{S}_b} \mathbbm{1}\{\hat{k}(\mathbf{x}_i) = k(\mathbf{x}_i)\}, \\
    \mathrm{conf}(\mathcal{S}_b) &:= \frac{1}{|\mathcal{S}_b|} 
    \sum_{i \in \mathcal{S}_b} \hat{c}(\mathbf{x}_i).
\end{align}
Bins with no assigned samples are omitted from subsequent analysis. Plotting $\mathrm{acc}(\mathcal{S}_b)$ against $\mathrm{conf}(\mathcal{S}_b)$ produces a reliability diagram (see left panel of Fig.~\ref{fig:rel_hist_diagram}).

In addition to reliability diagrams, we employ confidence histograms conditioned on correctness (see right panel of  Fig.~\ref{fig:rel_hist_diagram}). Predicted confidence scores are treated as samples of a random variable conditioned on whether the corresponding prediction is correct or incorrect, and separate histograms are constructed for each group. A decision-useful confidence score exhibits clear separation between these distributions, with correct predictions dominating the high-confidence regime and incorrect predictions concentrated at lower confidence values. Substantial overlap in the high-confidence region exposes failure modes that are not apparent from accuracy or loss alone.

Together, these diagnostics provide a transparent and interpretable assessment of whether a model’s confidence estimates meaningfully align with empirical correctness. In the following sections, we apply this framework across datasets, architectures, and training configurations, and use these insights to motivate an alternative ensemble-based Dirichlet construction for regimes in which direct evidential training becomes unstable.

\section{Methodology}\label{sec:mommethodology}

\subsection{A Stable Ensemble-Based Construction of Dirichlet Parameters}\label{sec:momalphas}

A wide range of EDL formulations have been proposed to improve uncertainty estimation and stabilize Dirichlet-based training, including alternative loss functions, regularization strategies, evidence parameterizations, and priors~\cite{chen2024reedl, haussmann2019pac, li2022tedl, pandey2023accumulate, yoon2024daedl, yu2024bevuncertainty}. While many show strong performance in specific settings, they often depend on pretrained representations, annealing schedules, or dataset-specific tuning. 

This fragility is especially problematic in practical settings with limited data and no pretrained weights, precisely where Dirichlet models are most appealing but least stable.  The failure mode illustrated in Fig.~\ref{fig:fail_demo} and discussed in Section~\ref{sec:momcalibration} exemplifies this behavior.  To address these issues, we propose a stable, ensemble-based construction of Dirichlet parameters grounded in the empirical variability of repeated softmax predictions. This approach avoids uncontrolled evidence inflation while preserving a probabilistic foundation for confidence estimation.

From a statistical perspective, the Dirichlet distribution is fully specified by its concentration parameters $\boldsymbol{\alpha}\in\mathbb{R}_{++}^K$ \cite{kotz2000dirichlet}.  To construct these parameters empirically, we begin with the analytical approach of deriving their maximum likelihood estimators (MLE).  Given a collection of independent and identically distributed samples $\{\boldsymbol{p}^{(m)} \}_{m=1}^M$ drawn from a Dirichlet-distributed random variable $\boldsymbol{p} \sim \mathrm{Dir}(\boldsymbol{\alpha})$, the log-likelihood of the concentration parameters $\boldsymbol{\alpha}$ is
\begin{equation}
    \mathrm{LL}(\boldsymbol{\alpha}) = M \log \Gamma(\alpha_0)
    - M \sum_{k=1}^K \log \Gamma(\alpha_k) + M \sum_{k=1}^K (\alpha_k - 1)\log \bar{p}_k,
\end{equation}
where $\log \bar{p}_k := \frac{1}{M}\sum_{m=1}^M \log p_k^{(m)}$.

We take the derivative and substitute in the digamma function in \eqref{eq:digammafunc} which yields for each $j=1,\dots,K$
\begin{align}
    \frac{\partial \mathrm{LL}}{\partial \alpha_j} &= M \psi(\alpha_0) - M \psi(\alpha_j) + M \log \bar{p}_j .
\end{align}
Dividing by $M$ and setting the derivative equal to zero gives
\begin{align}
    0 &= \psi(\alpha_0) - \psi(\alpha_j) + \log \bar{p}_j, \\
    \psi(\alpha_j) &= \psi(\alpha_0) + \log \bar{p}_j .
\end{align}

Because $\alpha_0$ depends on all $\{\alpha_j\}_{j=1}^K$, this system has no closed-form solution and must be solved using iterative numerical procedures such as fixed-point iteration or Newton methods, as discussed by Minka~\cite{minka2000estimating}. One commonly used fixed-point update takes the form
\begin{equation}\label{eq:fpi}
    \alpha^{(\text{new})}_j = \psi^{-1}\!\left[\psi\left(\alpha^{(\text{old})}_0\right)+\log \bar{p}_j\right].
\end{equation}
These updates require repeated evaluation of the digamma function and its inverse, and depend critically on a suitable initialization. Practical implementations initialize the iterative scheme using a method-of-moments estimate, which provides a stable and computationally inexpensive starting point for maximum-likelihood estimation.

We begin by introducing a moment-based estimator to initialize the Dirichlet parameters, and subsequently consider maximum likelihood estimation as an optional refinement of this initial estimate.  In our setting, the available samples are not drawn directly from a Dirichlet distribution, but arise from an ensemble of CE–trained classifiers. 

For a fixed input $\mathbf{x}_i$, we obtain $M$ softmax probability vectors $\{\boldsymbol{p}^{(m)}(\mathbf{x}_i)\}_{m=1}^M$, each corresponding to a different random initialization or training seed. These vectors lie on the probability simplex and are treated as empirical samples from an unknown Dirichlet distribution. For each class $k \in \{1,\dots,K\}$, we compute the empirical mean and unbiased variance across ensemble members,
\begin{equation}
    \hat{\mu}_k(\mathbf{x}_i) := \frac{1}{M} \sum_{m=1}^M \hat{p}_k^{(m)}(\mathbf{x}_i), \qquad 
    \hat{\sigma}_k^2(\mathbf{x}_i) := \frac{1}{M-1} \sum_{m=1}^M \bigl(\hat{p}_k^{(m)}(\mathbf{x}_i) - \hat{\mu}_k(\mathbf{x}_i)\bigr)^2.
\end{equation}
Matching these empirical moments to the theoretical moments of a Dirichlet distribution in \eqref{eq:meanvar} yields a class-wise estimate of the total concentration parameter,
\begin{equation}
    \hat{\alpha}_0^{(k)}(\mathbf{x}_i) = \frac{\hat{\mu}_k(\mathbf{x}_i) \bigl(1 - \hat{\mu}_k(\mathbf{x}_i)\bigr)}{\hat{\sigma}_k^2(\mathbf{x}_i)} - 1.
\end{equation}
To improve numerical stability, only finite and positive values of $\hat{\alpha}_0^{(k)}(\mathbf{x}_i)$ are retained, and the final estimate is obtained by averaging across valid classes,
\begin{equation}
    \hat{\alpha}_0(\mathbf{x}_i) := \frac{1}{|\mathcal{C}|} \sum_{c \in \mathcal{C}} \hat{\alpha}_0^{(c)}(\mathbf{x}_i),
\end{equation}
where $\mathcal{C} := \{k \in \{1,\dots,K\} \mid 0 < \hat{\alpha}_0^{(k)}(\mathbf{x}_i) < \infty\}$ denotes the set of classes with valid estimates. The individual Dirichlet parameters are then recovered via
\begin{equation}
    \hat{\alpha}_k(\mathbf{x}_i) = \hat{\mu}_k(\mathbf{x}_i)\; \hat{\alpha}_0(\mathbf{x}_i).
\end{equation}

Algorithm~\ref{alg:mom-dirichlet} summarizes this per-input method of moments construction. The resulting Dirichlet predictive mean remains constrained to the probability simplex and produces the same calibrated confidence interpretation as evidentially trained Dirichlet classifiers, while avoiding evidential loss design, activation choices, and unstable optimization dynamics.

Although the method of moments estimator provides a closed-form approximation, it is fundamentally a moment-matching procedure rather than a likelihood-maximizing one. Consequently, it may be viewed as a coarse estimate of the Dirichlet parameters. In contrast, maximum likelihood estimation directly optimizes the Dirichlet log-likelihood and is statistically efficient when the data are well modeled by a Dirichlet distribution. The method of moments estimate therefore provides a natural and stable initialization for optional likelihood-based refinement via fixed-point iteration.

Algorithm~\ref{alg:mle-dirichlet} describes this refinement step. Starting from the MoM initialization, we iteratively update the concentration parameters using Minka’s~\cite{minka2000estimating} fixed-point formulation in \eqref{eq:fpi}. This refinement is not required for constructing Dirichlet-based confidence estimates, but may improve parameter fidelity when the Dirichlet assumption is appropriate. We investigate this question empirically in Section~\ref{sec:exp2}.

\begin{algorithm}[!t]
    \caption{Dirichlet Parameter Estimation via Method of Moments (MoM)}
    \label{alg:mom-dirichlet}
    \begin{algorithmic}[1]
        \Require
        Input $\mathbf{x}_i$; ensemble size $M$; model trained with cross-entropy loss $\{f^{(m)}\}_{m=1}^M$
        \Ensure
        Dirichlet parameters $\hat{\boldsymbol{\alpha}}(\mathbf{x}_i)$
        
        \State Obtain ensemble softmax scores $\{\hat{\boldsymbol{p}}^{(m)}(\mathbf{x}_i)\}_{m=1}^M$
        \State Compute per-class empirical moments $\{\hat{\mu}_k(\mathbf{x}_i), \hat{\sigma}_k^2(\mathbf{x}_i)\}_{k=1}^K$
        \State Compute class-wise $\alpha_0$ estimates
        $\hat{\alpha}_0^{(k)} \gets \hat{\mu}_k(\mathbf{x}_i)(1-\hat{\mu}_k(\mathbf{x}_i))/\hat{\sigma}_k^2(\mathbf{x}_i) - 1$
        \State Define valid class set $\mathcal{C}=\{k \mid 0<\hat{\alpha}_0^{(k)}<\infty\}$
        \State Aggregate total concentration
        $\hat{\alpha}_0(\mathbf{x}_i) \gets \left(1/|\mathcal{C}|\right)\cdot\sum_{k\in\mathcal{C}}\hat{\alpha}_0^{(k)}$
        \State Recover Dirichlet parameters
        $\hat{\alpha}_k(\mathbf{x}_i) \gets \hat{\mu}_k(\mathbf{x}_i)\hat{\alpha}_0(\mathbf{x}_i)$ for $k=1,\dots,K$
        \State \Return $\hat{\boldsymbol{\alpha}}(\mathbf{x}_i)$ 
    \end{algorithmic}
\end{algorithm}

\begin{algorithm}[!t]
    \caption{Dirichlet Parameter Estimation Refinement via Fixed-Point MLE}
    \label{alg:mle-dirichlet}
    \begin{algorithmic}[1]
        \Require
        Input $\mathbf{x}_i$; ensemble softmax scores $\{\hat{\boldsymbol{p}}^{(m)}(\mathbf{x}_i)\}_{m=1}^M$; MoM initialization $\hat{\boldsymbol{\alpha}}^{\text{MoM}}(\mathbf{x}_i)$; max iterations $T$
        \Ensure
        Refined Dirichlet parameters $\hat{\boldsymbol{\alpha}}(\mathbf{x}_i)$
        
        \State Compute mean log-probabilities
        $\bar{\ell}_k(\mathbf{x}_i) \gets (1/M)\sum_{m=1}^M \log \hat{p}_k^{(m)}(\mathbf{x}_i)$
        \State Initialize
        $\boldsymbol{\alpha}^{(0)} \gets \hat{\boldsymbol{\alpha}}^{\text{MoM}}(\mathbf{x}_i)$
        \For{$t=0$ to $T-1$}
            \State $\alpha_0^{(t)} \gets \sum_{k=1}^K \alpha_k^{(t)}$
            \State $\alpha_k^{(t+1)} \gets \psi^{-1}\!\big(\psi(\alpha_0^{(t)}) + \bar{\ell}_k(\mathbf{x}_i)\big)$ for $k=1,\dots,K$
            \If{$\|\boldsymbol{\alpha}^{(t+1)}-\boldsymbol{\alpha}^{(t)}\| < \varepsilon \|\boldsymbol{\alpha}^{(t)}\|$}
                \State \textbf{break}
            \EndIf
        \EndFor
        \State $\hat{\boldsymbol{\alpha}}(\mathbf{x}_i) \gets \boldsymbol{\alpha}^{(t+1)}$
        \State \Return $\hat{\boldsymbol{\alpha}}(\mathbf{x}_i)$
    \end{algorithmic}
\end{algorithm}

\subsection{Variance-Based Selective Classification}

Selective classification allows a model to abstain on uncertain inputs in order to reduce error on retained predictions \cite{pugnana2024benchmarks, geifman2017selective}.  A selective classifier operates by ranking predictions according to a confidence or uncertainty score and choosing an operating point that satisfies a desired risk constraint.  Within the EDL framework, Dirichlet-based classifiers provide predictive uncertainty directly through their concentration parameters. However, the explicit use of Dirichlet predictive variance as an ordering statistic for risk-controlled abstention has not been systematically studied. 

In this work, we propose a variance-based selective classification strategy based on Dirichlet predictive variance.  Specifically, we use the total predictive variance to rank samples by uncertainty and select a risk-constrained operating point via calibration. This strategy relies on the assumption that predictive variance meaningfully varies across inputs. However, under certain evidential training dynamics, the Dirichlet parameters can enter a degenerate regime in which this assumption fails. 

If we consider the EDL failure regime demonstrated in Fig.~\ref{fig:fail_demo} of Section~\ref{sec:momcalibration}, we observe a characteristic collapse pattern in which the predictive mean becomes uniform while the total concentration grows arbitrarily large, i.e.,
\begin{equation}
    \hat p_k = \frac{1}{K} \quad \forall k, \qquad \alpha_0 \to \infty.
\end{equation}
Since the Dirichlet predictive mean satisfies $\mathbb{E}[p_k\mid \boldsymbol{\alpha} ]=\alpha_k/\alpha_0$, uniform predictions imply 
\begin{equation}\label{eq:alphakfail}
    \frac{\alpha_k}{\alpha_0} = \frac{1}{K} \quad \Longrightarrow \quad \alpha_k = \frac{\alpha_0}{K} \quad \forall k.
\end{equation}

Substituting this relationship into the class-wise predictive variance from \eqref{eq:meanvar} produces 
\begin{align}
    \mathbb{V}[p_k \mid \boldsymbol{\alpha}] &= \frac{\left(\frac{\alpha_0}{K}\right)
    \left(\alpha_0 - \frac{\alpha_0}{K}\right)}
    {\alpha_0^2(\alpha_0+1)} = \frac{K-1}{K^2(\alpha_0+1)} \quad \forall k.
\end{align}

Two consequences follow immediately: 
\begin{enumerate}
    \item All classes have identical predictive variance.
    \item As $\alpha_0 \rightarrow \infty$, $\mathbb{V}[\mathbf{p} \mid \boldsymbol{\alpha}] \rightarrow 0 \quad \forall k$.
\end{enumerate}

\begin{figure}[t]
    \centering
    \begin{minipage}{0.49\linewidth}
        \centering
        \includegraphics[width=\linewidth]{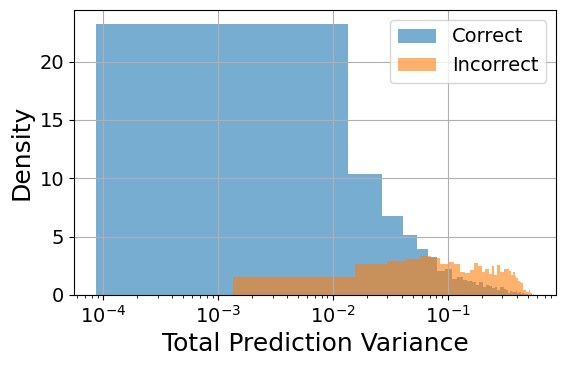}
    \end{minipage}
    \hfill
    \begin{minipage}{0.49\linewidth}
        \centering
        \includegraphics[width=\linewidth]{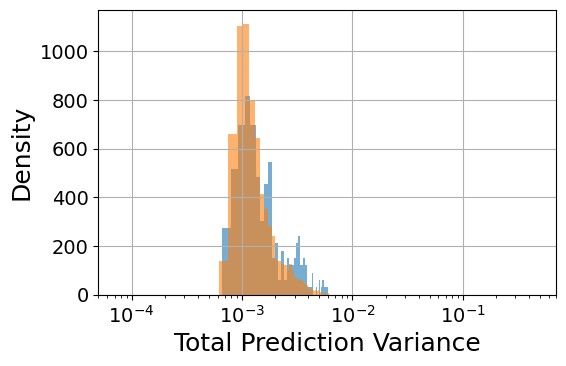}
    \end{minipage}
    \caption[Variance Histograms for Successful Versus Collapsed Training]{Comparison of successful versus failed EDL training variance histograms for correct and incorrect predictions. Successful training (left) shows clearer separation between the distributions, enabling selective classification thresholding, while failed training (right) shows substantial overlap within a condensed variance range.}
    \label{fig:failvariance}
\end{figure}

Thus, every prediction whether correct or incorrect shares the same vanishing variance value.  Because the total predictive variance is defined as 
\begin{equation}\label{eq:totalvar}
    \mathbb{V}[\mathbf{p} \mid \boldsymbol{\alpha}]
    = \sum_{k=1}^K \mathbb{V}[p_k \mid \boldsymbol{\alpha}],
\end{equation}
and each class-wise variance converges to zero at rate $\mathcal{O}(1/\alpha_0)$, it follows that the total variance is also identical across samples and converges to the same limit at the same rate.  Therefore, in the uniform-collapse regime, predictive variance becomes a degenerate signal; it assigns the same (vanishing) uncertainty to all inputs.  No ordering of samples is possible, and variance-based selective classification necessarily fails.  This theoretical behavior is reflected empirically in Fig.~\ref{fig:failvariance}, where correct and incorrect predictions exhibit almost complete overlap in their total variance distributions under a near-uniform collapse regime, whereas the successfully trained model shows clear separation.

In contrast, classifiers trained with the standard cross-entropy loss function produce logits $\mathbf{z}$ which are transformed into class probabilities $\hat{\mathbf{p}}$ via the softmax mapping (see \eqref{eq:logits}\textendash \eqref{eq:softmax}). Training is performed using the cross-entropy loss functin defined in \eqref{eq:celoss}.  For a single input with one-hot label $\mathbf{y}$ and true class index $j$, the gradient of the cross-entropy loss with respect to each logit $z_k$ satisfies
\begin{equation}
    \frac{\partial \mathcal{L}_{\mathrm{CE}}}{\partial z_k}
    = \hat{p}_k - y_k.
\end{equation}

Since $\hat{p}_k \in [0,1]$ and $y_k \in \{0,1\}$, it follows directly that the gradient is bounded for every class and sample:
\begin{equation}
    \frac{\partial \mathcal{L}_{\mathrm{CE}}}{\partial z_k} \in [-1,1]     \quad \forall k.
\end{equation}

When $k \neq j$, the gradient equals $\hat{p}_k$ and drives predicted probabilities toward zero; when $k=j$, it equals $\hat{p}_j - 1$ and drives $\hat{p}_j \rightarrow 1$. The gradient vanishes only in the limiting cases $\hat{p}_k=0$ or $\hat{p}_j=1$, and the model is not updated. Thus, unlike evidential objectives, cross-entropy contains no coupled concentration term analogous to $\alpha_0$, so optimization is governed by bounded error signals rather than runaway scaling dynamics (more detailed gradient analysis for both evidential and CE training can be in \citet{pandey2025gen}).

\begin{figure}[t]
    \centering
    \includegraphics[width=0.8\textwidth]{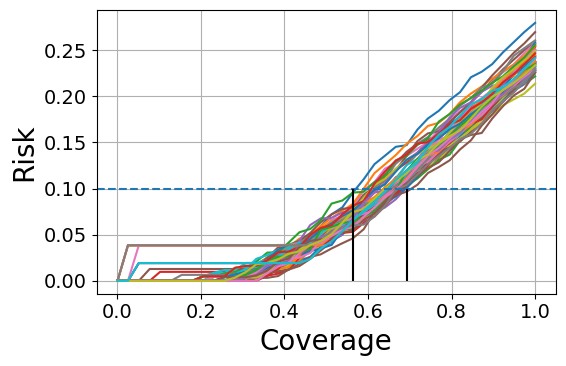}
    \caption[Risk–Coverage Curves for 50 Independently Trained Classifiers]{Risk–coverage curves for 50 independently initialized classifiers trained with cross-entropy loss on a medical dataset. Each curve corresponds to a separately initialized model. The variability in the low-coverage regime illustrates instability in confidence ordering across CE training runs. At a fixed target risk, $r$ $=$$0.1$, the achieved coverage ranges from $56.39\%$ to $69.26\%$, yielding a spread of $12.87$ percentage points, further quantifying this variability.}
    \label{fig:50CE_riskcov_curves}
\end{figure}

While this suggests that the error signal from the gradient remains well-behaved across classes and training regimes, CE does not provide an intrinsic second-order uncertainty representation.  Even when post-hoc uncertainty measures are constructed from CE outputs, the resulting ordering can vary substantially across independent training runs. This instability is illustrated in Fig.~\ref{fig:50CE_riskcov_curves}, which shows risk\textendash coverage curves for 50 independently trained classifiers trained with cross-entropy loss where uncertainty is quantified using Dirichlet total variance obtained via method of moments estimation from each model’s softmax outputs. The spread among curves, particularly in the low-coverage regime, indicates that the confidence ranking induced by a single trained classifier is sensitive to initialization and optimization dynamics.

In this work, we propose a variance-based selective classification strategy using the total predictive variance of a Dirichlet distribution. Samples are ranked according to predictive variance, and a risk-constrained operating point is selected via calibration. This approach assumes that predictive variance differs meaningfully across inputs and provides a stable ordering for abstention. However, as we demonstrated in Fig.~\ref{fig:failvariance}, certain evidential training dynamics can drive the Dirichlet parameters into degenerate regimes in which variance collapses, undermining this assumption.

That is where our ensemble-based method of moments construction introduced in Section~\ref{sec:momalphas} provides a principled extension. Rather than directly optimizing concentration parameters, we estimate them from empirical moments of repeated softmax probability vectors $\hat{\mathbf{p}}$ obtained from independently initialized classifiers trained with cross-entropy loss under the bounded-gradient dynamics described above. Because each $\hat{\mathbf{p}}$ arises from \eqref{eq:softmax} under stable optimization, the resulting empirical variability is finite and varies with the input $\mathbf{x}$, reflecting differences in model predictions across training runs. The inferred concentration parameters therefore encode ensemble disagreement, producing a Dirichlet representation whose total concentration reflects observed variability rather than unconstrained evidence growth.  Consequently, the ensemble-based construction retains the bounded probability dynamics of cross-entropy training while introducing a Dirichlet parameterization whose concentration is determined by empirical variability rather than fragile evidence accumulation mechanisms.

To construct the variance-based selective classification methodology, a scalar abstention score is computed via the total predictive variance from \eqref{eq:totalvar} for each input $\mathbf{x}_i$, defined as $\mathbb{V}(\mathbf{x}_i):=\mathbb{V}[\mathbf{p} \mid \boldsymbol{\alpha}(\mathbf{x}_i)]$. Predictions are formed using the Dirichlet mean, while abstention decisions depend solely on $\mathbb{V}(\mathbf{x}_i)$.  To select the decision threshold, a variance threshold $\tau$ is determined on a held-out calibration set. Calibration samples are sorted by increasing $\mathbb{V}(\mathbf{x}_i)$, and $\tau$ is chosen such that the empirical risk on the retained calibration subset is approximately a target level $r$ (in our experiments, $r = 0.1$). The selected threshold is then fixed and applied unchanged to the test set. At inference time, the decision rule for each sample $\mathbf{x}_i$ is
\begin{equation}
    \mathbf{\hat{y}}(\mathbf{x}_i) =
        \begin{cases}
        \arg\max_k \mathbb{E}[p_k \mid \boldsymbol{\alpha}(\mathbf{x}_i)], & \text{if } \mathbb{V}(\mathbf{x}_i) \le \tau, \\
        \text{abstain},  & \text{if } \mathbb{V}(\mathbf{x}_i) > \tau.
        \end{cases}
\end{equation}

To analyze the behavior of this ordering statistic, we construct risk–coverage curves as shown in Fig.~\ref{fig:50CE_riskcov_curves} and overlaid variance histograms of $\mathbb{V}(\mathbf{x}_i)$ for correct and incorrect test predictions as shown in Fig.~\ref{fig:failvariance}.  Because total predictive variance values are often very small and can differ by several orders of magnitude, the horizontal axis is displayed on a logarithmic scale to make separation between inputs visually discernible.  We evaluate this variance-based ordering with ensemble-based and standard EDL Dirichlet constructions in the subsequent experimental study.

\section{Experimental Results}\label{sec:momresults}

This section presents three experiments. Section~\ref{sec:exp1} analyzes how evidential design choices affect confidence behavior. Section~\ref{sec:exp2} evaluates the proposed method of moments construction with optional maximum likelihood refinement and compares it to EDL configurations from Section~\ref{sec:exp1}. Section~\ref{sec:exp3} examines selective classification using Dirichlet variance on a medical dataset.

\subsection{Experiment 1: Calibration Diagnostics Under Existing Dirichlet Formulation Choices}\label{sec:exp1}

This experiment evaluates how formulation choices in Dirichlet-based classifiers affect predictive confidence. Using the diagnostic tools from Section~\ref{sec:momcalibration}, we examine the impact of additive priors, regularization, and evidence parameterizations while holding backbone and optimization fixed.

\subsubsection{Experimental Setup for EDL Calibration Analysis}

We evaluate the confidence calibration of six EDL configurations summarized in Table~\ref{tab:dirichlet_formulations} across four image classification datasets of increasing complexity and domain specificity: CIFAR-10, CIFAR-100~\cite{krizhevsky2009learn}, DermaMNIST~\cite{medmnistv2}, and the APTOS 2019 Blindness Detection dataset~\cite{aptos2019blindness}. Basic dataset characteristics, including sample counts and class counts, are summarized in Table~\ref{tab:dataset_summary}. CIFAR datasets are used for benchmarking across low- and high-class-count regimes, while DermaMNIST and APTOS are used to demonstrate efficacy in real-world, high-stakes medical settings characterized by limited data, class imbalance, and elevated decision risk.

Each dataset is re-partitioned using stratified sampling into training ($60\%$), validation ($20\%$), calibration ($10\%$), and test ($10\%$) splits. The validation set is used for model selection and training diagnostics, and the calibration set is reserved for computing post-training uncertainty metrics without contaminating the test evaluation. The calibration set is used only for auxiliary procedures (e.g., temperature scaling) and is excluded from training. Reliability diagrams are constructed using $B=10$ equal-width confidence bins over $[0,1]$, and this choice is fixed across all experiments. All images are resized to $224\times224$ pixels. Training images are augmented with random crops and horizontal flips; all others are resized and normalized using ImageNet~\cite{deng2009imagenet} statistics. Grayscale DermaMNIST images are converted to RGB to match the expected input format.

All models use an ImageNet-pretrained ResNet-18~\cite{he2016deep} with a task-specific output head; no layers are frozen. Optimization is performed with Adam~\cite{kingma2015adam} using dual learning rates ($1\times10^{-4}$ for the backbone, $1\times10^{-3}$ for the head) and shared weight decay of $1\times10^{-4}$. Unless stated otherwise, models are trained for 30 epochs with batch size 32. For the variability analysis in this section, each configuration is repeated across 10 random seeds, and results are reported as mean $\pm$ standard deviation. These repetitions are used only to illustrate variability in model behavior and are not part of the proposed method.

\begin{table}[!t]
\centering
\caption[Summary of EDL Formulations Evaluated in Experiments]{Summary of EDL formulations used for comparison.}
\label{tab:dirichlet_formulations}
\begin{tabular}{lcccc}
\hline
\textbf{Model} &
\textbf{Base Loss} &
\textbf{Regularization} &
\textbf{Activation} &
\textbf{$\boldsymbol{\alpha}$} \\
\hline
MSE Only & MSE & None & Softplus & $\boldsymbol{\alpha} >1e-6$  \\

MSE Clamp & MSE  & KL (annealed) & Softplus &  $\boldsymbol{\alpha} >1e-6$ \\

MSE Soft Adapt & MSE & KL (annealed) & Adaptive Softplus & $\boldsymbol{\alpha} >1e-6$  \\

MSE Plus One & MSE & KL (annealed) & Softplus & $\boldsymbol{\alpha} = \mathbf{e} + \mathbf{1}$ \\

Exponential & MSE & KL (annealed) & Exponential & $\boldsymbol{\alpha} = \mathbf{e} + \mathbf{1}$ \\

Digamma & Digamma & Log-evidence & Softplus & $\boldsymbol{\alpha} = \mathbf{e} + \mathbf{1}$ \\
\hline
\end{tabular}
\end{table}

\begin{table}[t]
\centering
\caption[Summary of Datasets Used in the Experiments]{Summary of datasets used in the experiments.}
\label{tab:dataset_summary}
\begin{tabular}{lccc}
\toprule
Dataset & Domain & Samples & Classes \\
\midrule
CIFAR-10 & Natural images & 60,000 & 10 \\
CIFAR-100 & Natural images & 60,000 & 100 \\
DermaMNIST & Dermatology images & 10,015 & 7 \\
APTOS 2019 & Retinal fundus images & 3,662 & 5 \\
\bottomrule
\end{tabular}
\end{table}

\subsubsection{Results for EDL Calibration Analysis}

We assess the calibration behavior of six EDL configurations across four datasets, with primary results for CIFAR-100 and DermaMNIST discussed in the Section \ref{sec:exp1}. These datasets represent distinct challenges: CIFAR-100 evaluates performance under a high class count, while DermaMNIST presents a small-scale medical imaging task with limited training data. CIFAR-10 and APTOS results are provided in Appendix~\ref{sec:append_exp1} to support broader generalization. In addition to standard predictive metrics, we report confidence histograms and reliability diagrams to examine how formulation choices affect confidence behavior beyond accuracy alone.

\begin{figure}[t]
    \centering
    \includegraphics[width=\textwidth]{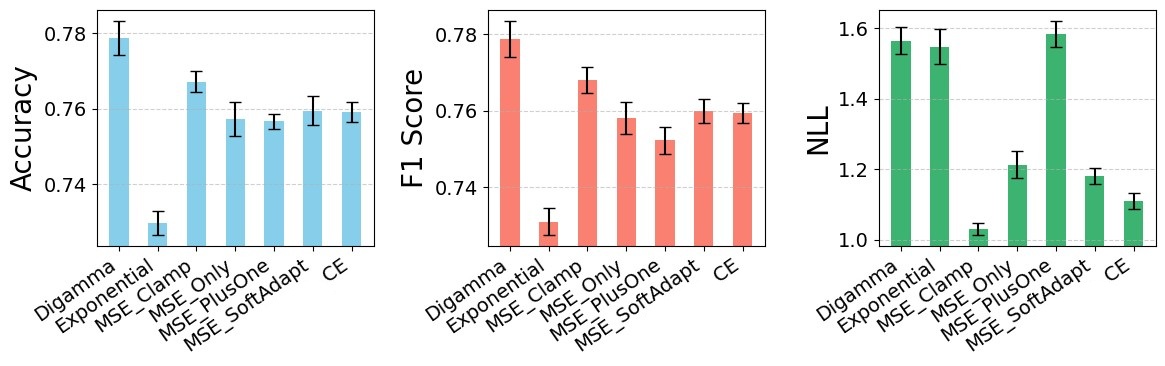}
    \caption[Predictive Performance Metrics for EDL on CIFAR-100]{Predictive performance metrics (accuracy, macro F1, and negative log-likelihood) for the evaluated EDL configurations on CIFAR-100.}
    \label{fig:C100_EDL_metrics}
\end{figure}

\begin{figure}[t]
    \centering
    \begin{subfigure}{0.5\linewidth}
        \includegraphics[width=\textwidth]{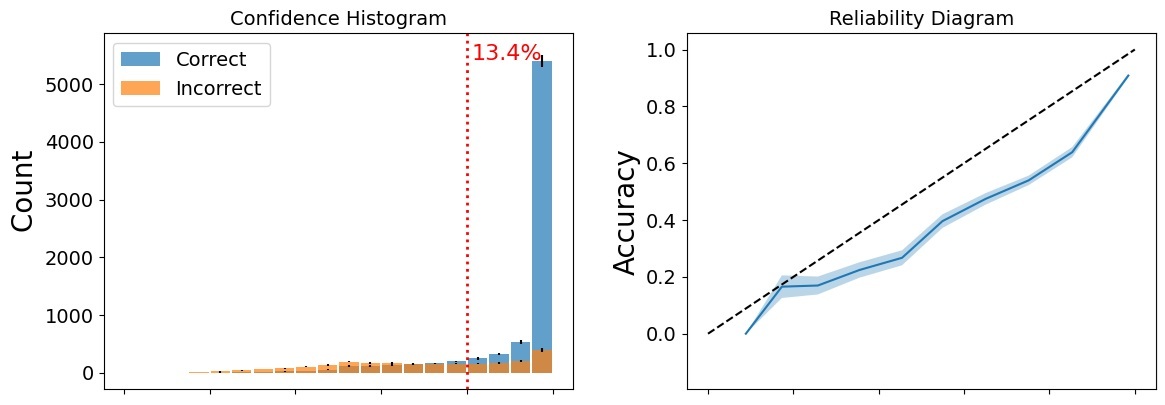}
        \caption{MSE Only}
        \label{fig:c100_mse_only}
    \end{subfigure}\hfill
    \begin{subfigure}{0.5\linewidth}
        \includegraphics[width=\textwidth]{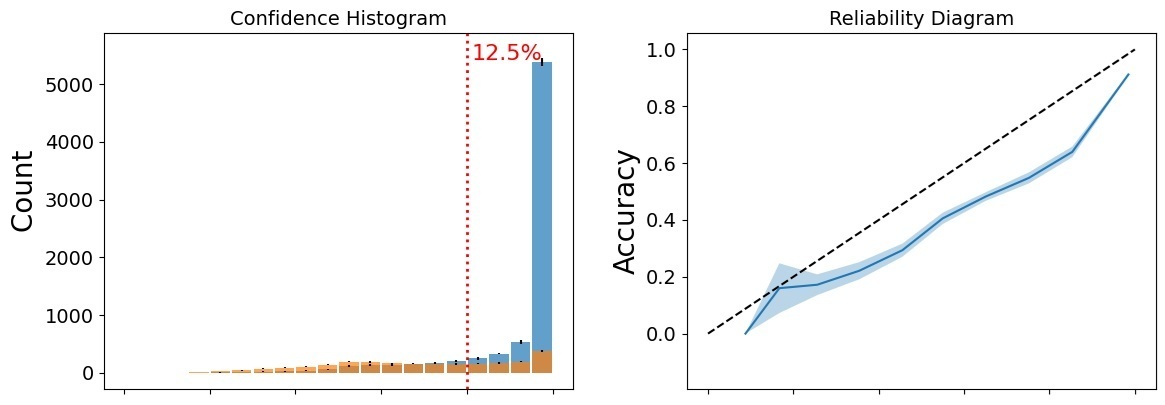}
        \caption{MSE SoftAdapt}
        \label{fig:c100_mse_softadapt}
    \end{subfigure}
    
    \begin{subfigure}{0.5\linewidth}
        \includegraphics[width=\textwidth]{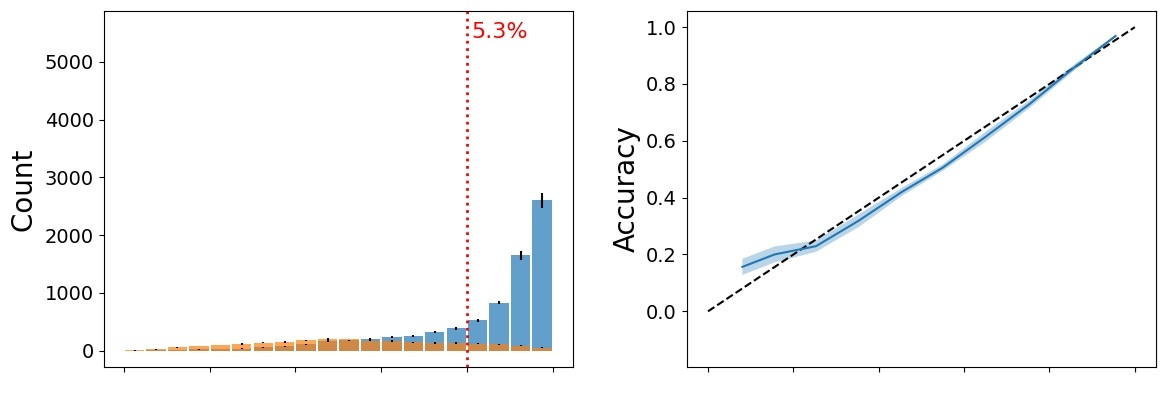}
        \caption{MSE Clamp}
        \label{fig:c100_mse_clamp}
    \end{subfigure}\hfill
    \begin{subfigure}{0.5\linewidth}
        \includegraphics[width=\textwidth]{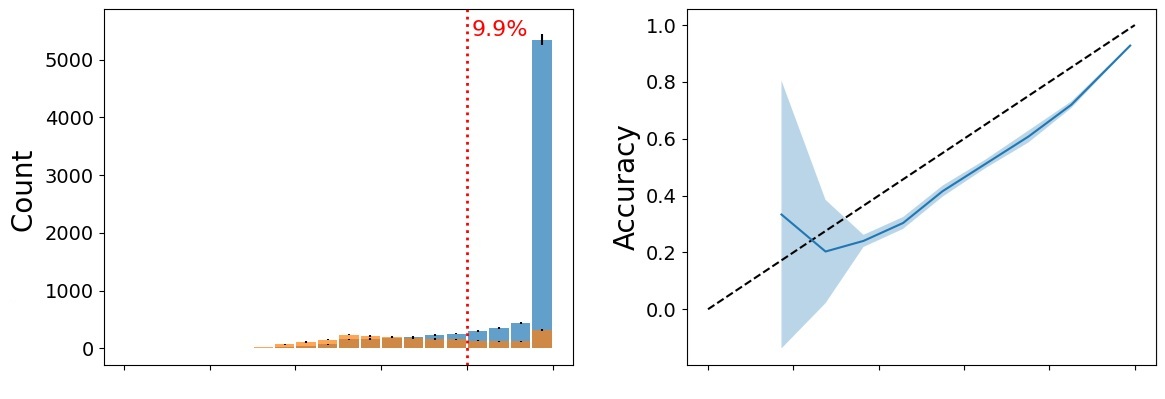}
        \caption{Digamma}
        \label{fig:c100_digamma}
    \end{subfigure}
    
    \begin{subfigure}{0.5\linewidth}
        \includegraphics[width=\textwidth]{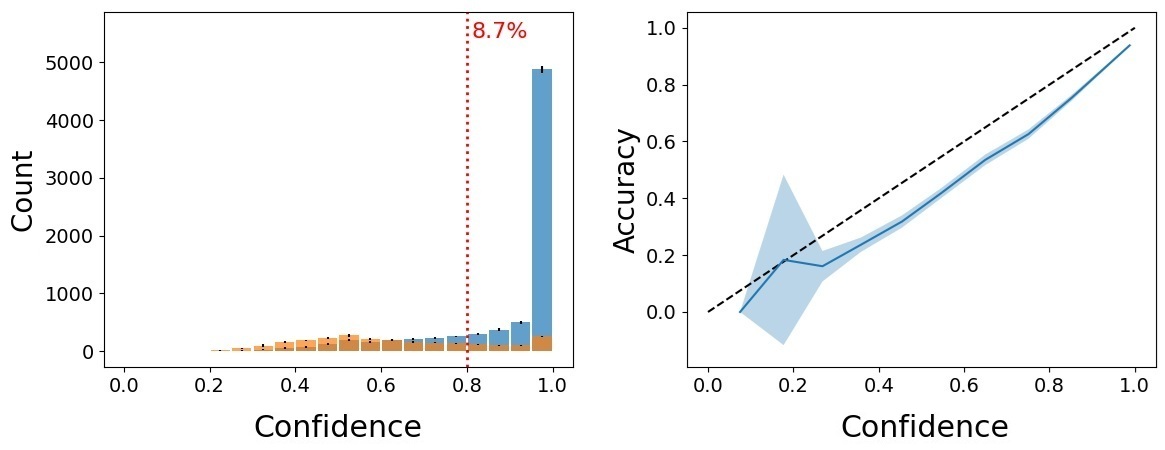}
        \caption{MSE PlusOne}
        \label{fig:c100_mse_plusone}
    \end{subfigure}\hfill
    \begin{subfigure}{0.5\linewidth}
        \includegraphics[width=\textwidth]{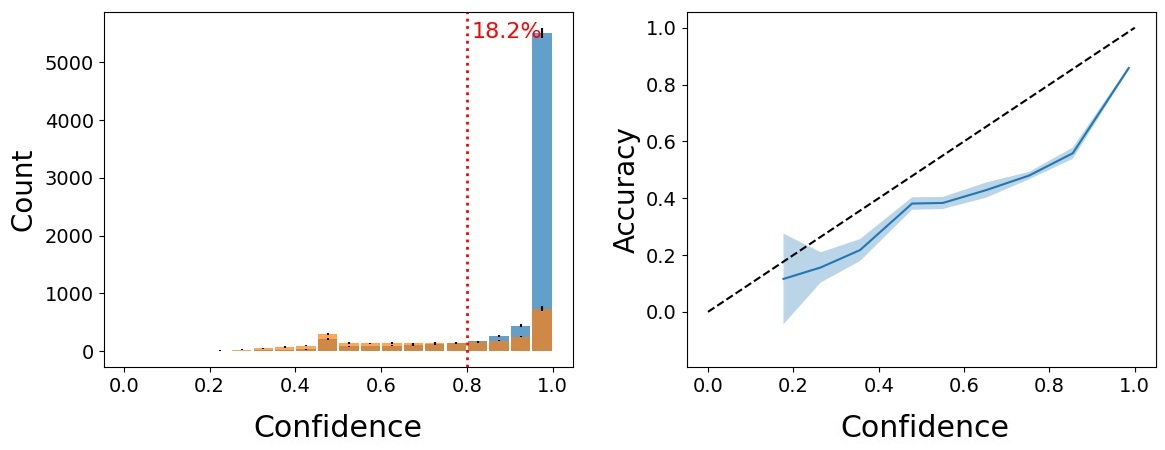}
        \caption{Exponential}
        \label{fig:c100_exponential}
    \end{subfigure}
    
    \caption[Confidence Calibration Diagnostics of EDL on CIFAR-100]{Confidence diagnostics for Dirichlet classifiers on CIFAR-100. Each subfigure corresponds to a different training configuration where the left panel shows the confidence histogram and the right panel shows the reliability diagram. Confidence histograms are computed from the mean predictions across runs, and the red annotation reports the percentage of incorrect predictions with confidence greater than $0.8$. Reliability diagrams summarize calibration behavior across runs, where shaded bands denote one standard deviation and the dashed line indicates perfect calibration.}
    \label{fig:c100_EDL_sum}
\end{figure}

\begin{figure}[t]
    \centering
    \includegraphics[width=\textwidth]{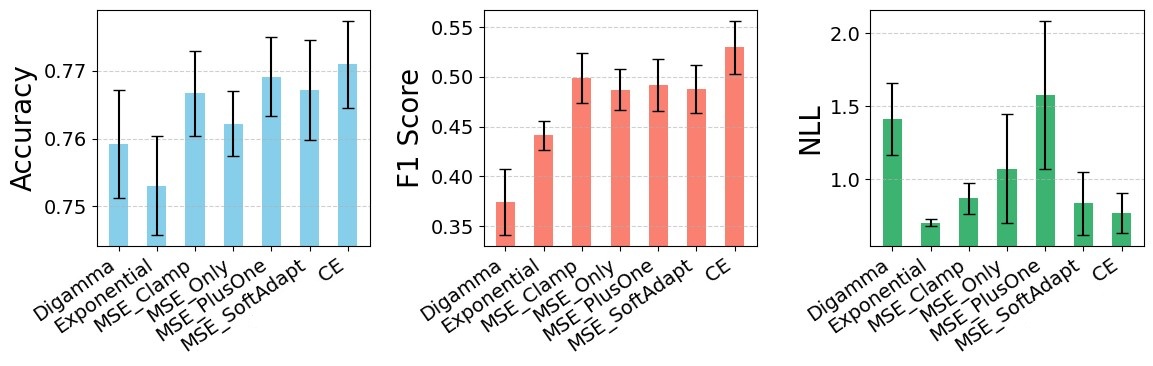}
    \caption[Predictive Performance Metrics for EDL on DermaMNIST]{Predictive performance metrics (accuracy, macro F1, and negative log-likelihood) for the evaluated EDL configurations on DermaMNIST.}
    \label{fig:DM_EDL_metrics}
\end{figure}

\begin{figure}[t]
    \centering
    \begin{subfigure}{0.5\linewidth}
        \includegraphics[width=\textwidth]{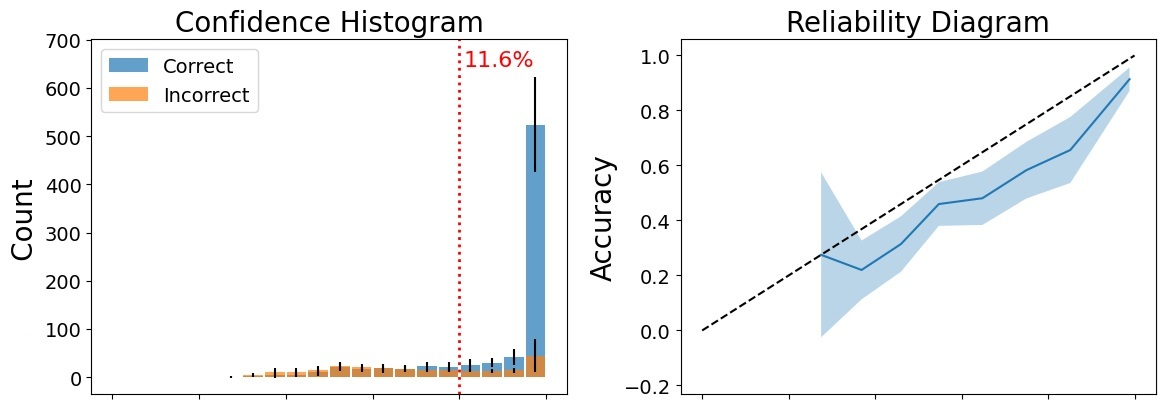}
        \caption{MSE Only}
        \label{fig:dm_mse_only}
    \end{subfigure}\hfill
    \begin{subfigure}{0.5\linewidth}
        \includegraphics[width=\textwidth]{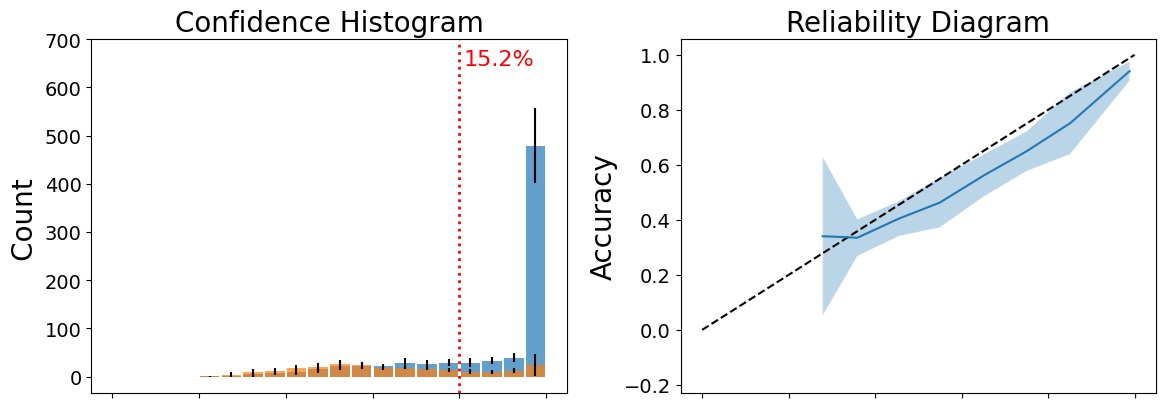}
        \caption{MSE SoftAdapt}
        \label{fig:dm_mse_softadapt}
    \end{subfigure}
    
    \begin{subfigure}{0.5\linewidth}
        \includegraphics[width=\textwidth]{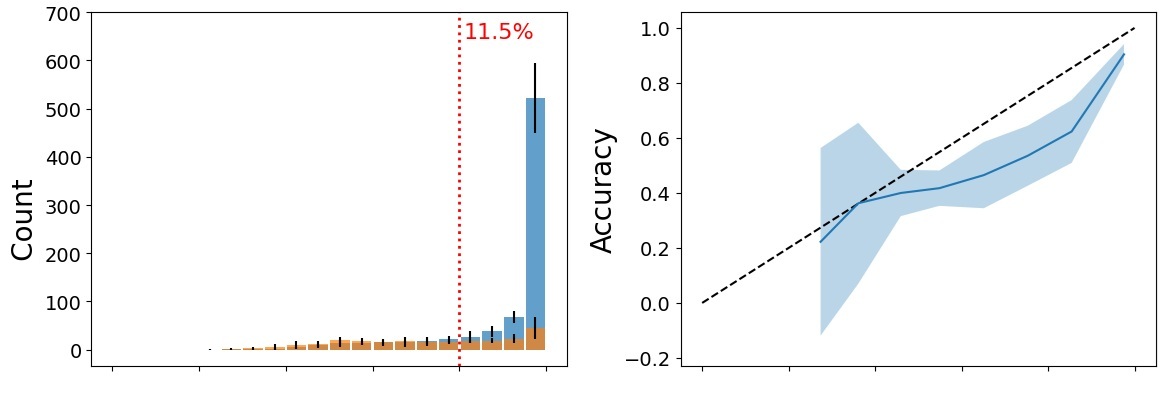}
        \caption{MSE Clamp}
        \label{fig:dm_mse_clamp}
    \end{subfigure}\hfill
    \begin{subfigure}{0.5\linewidth}
        \includegraphics[width=\textwidth]{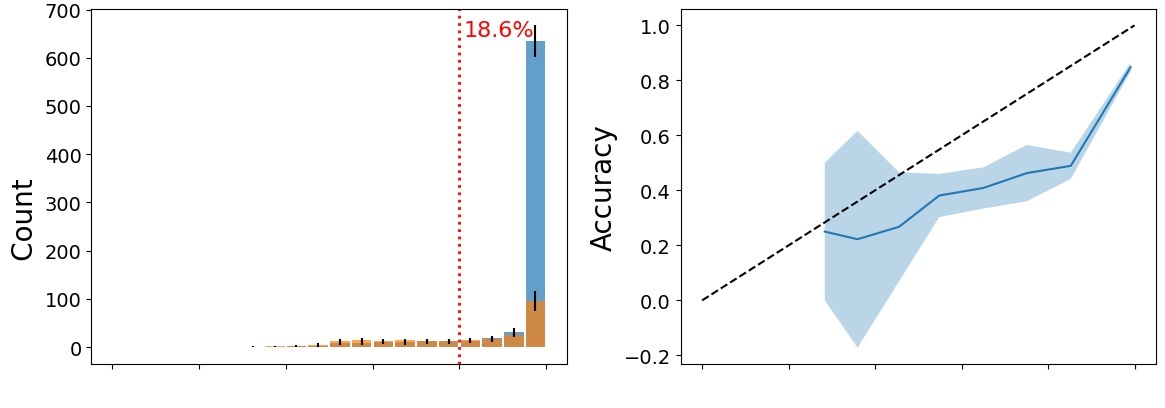}
        \caption{Digamma}
        \label{fig:dm_digamma}
    \end{subfigure}
    
    \begin{subfigure}{0.5\linewidth}
        \includegraphics[width=\textwidth]{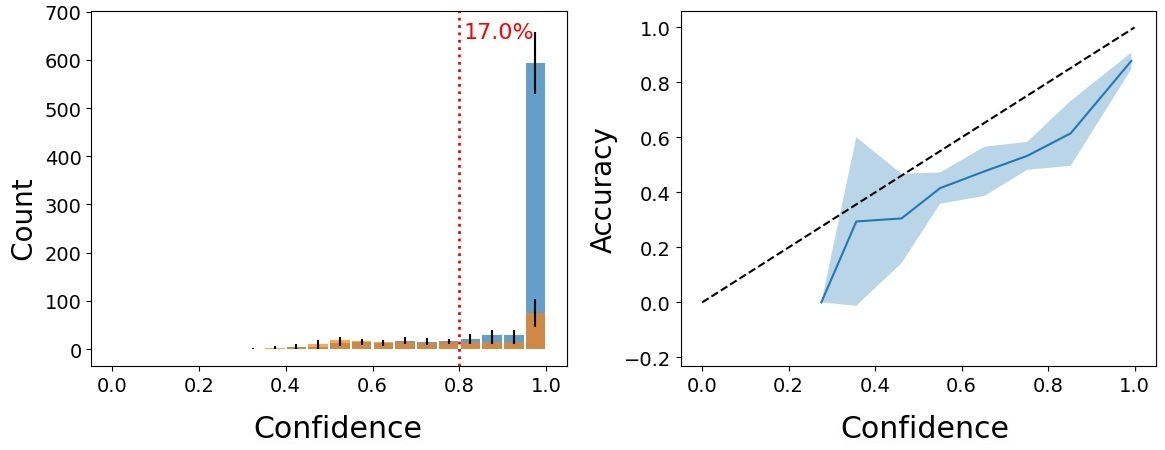}
        \caption{MSE PlusOne}
        \label{fig:dm_mse_plusone}
    \end{subfigure}\hfill
    \begin{subfigure}{0.5\linewidth}
        \includegraphics[width=\textwidth]{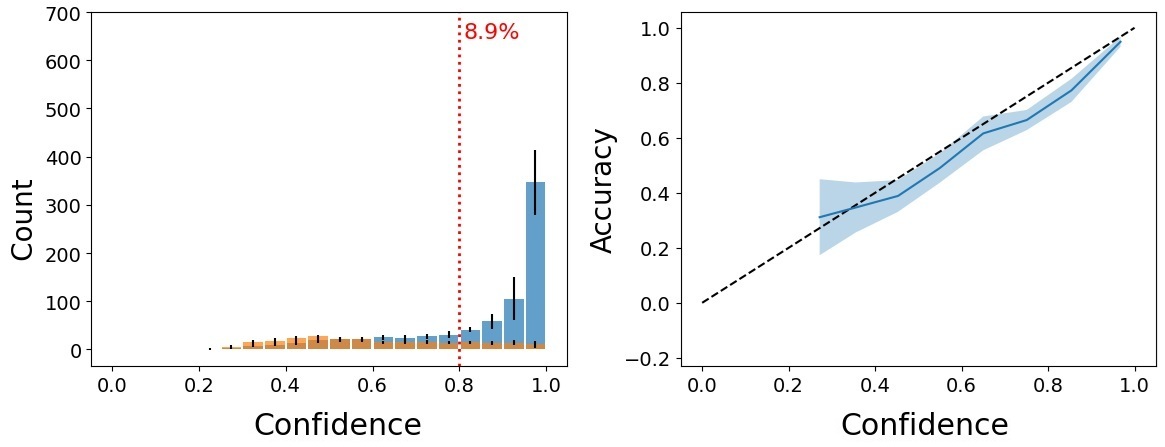}
        \caption{Exponential}
        \label{fig:dm_exponential}
    \end{subfigure}
    
    \caption[Confidence Calibration Diagnostics of EDL on DermaMNIST]{Confidence diagnostics for Dirichlet classifiers on DermaMNIST. Each subfigure corresponds to a different training configuration where the left panel shows the confidence histogram and the right panel shows the reliability diagram. Confidence histograms are computed from the mean predictions across runs, and the red annotation reports the percentage of incorrect predictions with confidence greater than $0.8$. Reliability diagrams summarize calibration behavior across runs, where shaded bands denote one standard deviation and the dashed line indicates perfect calibration.}
    \label{fig:DM_EDL_sum}
\end{figure}

Figs.~\ref{fig:C100_EDL_metrics}\textendash~\ref{fig:c100_EDL_sum} and
\ref{fig:DM_EDL_metrics}\textendash~\ref{fig:DM_EDL_sum} summarize the results for CIFAR-100 and DermaMNIST, respectively. For CIFAR-100 predictive performance metrics in Fig.~\ref{fig:C100_EDL_metrics}, most EDL variants achieve comparable accuracy, with the exception of the Exponential configuration. A similar pattern is observed in the F1 score. Negative log-likelihood (NLL) shows the largest separation, with configurations that exclude the $\delta=1$ additive prior generally achieving lower values.

On DermaMNIST performance metrics in Fig.~\ref{fig:DM_EDL_metrics}, the pattern differs. Digamma and Exponential yield the lowest accuracy and F1 scores, while Exponential produces among the lowest and most consistent negative log-likelihood values. In contrast, Digamma and MSE with $\delta=1$ exhibit the highest and most unstable negative log-likelihood.

Turning to calibration diagnostics, Figs.~\ref{fig:c100_EDL_sum} and~\ref{fig:DM_EDL_sum} present confidence histograms and reliability diagrams for each variant across 10 independent training runs. Confidence histograms are computed using the mean predicted confidence across runs, while reliability diagrams are reported as mean $\pm$ standard deviation. The histograms include a red threshold line at confidence $0.8$, with the percentage of incorrect predictions above this threshold reported to quantify high-confidence error.

Reviewing the confidence histograms for CIFAR-100, the MSE with clamped alphas configuration yields the lowest proportion of incorrect predictions in the high-confidence region above $0.8$, at $5.3\%$, whereas the Exponential model performs worst at $18.2\%$. This pattern is reflected in the reliability diagrams: MSE with alpha clamping closely follows the diagonal reference line, indicating near-calibrated behavior, while the Exponential configuration falls substantially below the diagonal, consistent with overconfidence.

The MSE with $\delta=1$ additive prior and Digamma variants exhibit instability in the low-confidence regime, visible in the left tail of the histograms and corresponding fluctuations in the reliability diagrams, suggesting inconsistent behavior for uncertain predictions.

For the DermaMNIST calibration diagnostics, the Exponential now performs better with the lowest incorrect percentage for high-confidence scores with the distribution of correct versus incorrect showing the best separation while the reliability diagram demonstrates a good pairing with the perfect calibration line and little instability in training runs.  The worst performer is confirmed to be the Digamma variant which both possesses the highest incorrect percentage of high-confidence scores as well as the greatest deviation from perfect calibration and instability.

For our overall analysis across all datasets (including those in the appendices), predictive accuracy alone obscures meaningful differences in confidence behavior. On CIFAR-10 and CIFAR-100, most Dirichlet configurations achieve comparable accuracy and F1 scores, yet separate more clearly in negative log-likelihood and calibration diagnostics. The Exponential formulation exhibits the weakest behavior under high class count, with elevated high-confidence error and visible overconfidence on CIFAR-100. In contrast, the MSE with SoftPlus variants, particularly with alpha clamping, tend to produce more stable calibration patterns in natural image settings, though differences remain modest at the level of aggregate accuracy.

Under the small data, medical imaging conditions, performance behavior shifts. On DermaMNIST, Exponential demonstrates comparatively strong calibration separation and stable likelihood behavior, while Digamma and MSE with $\delta=1$ show greater instability in NLL and confidence diagnostics. The APTOS results further highlight this sensitivity: mean accuracy remains similar across variants, but Digamma displays elevated NLL and increased run-to-run variability in high-confidence errors. Importantly, no single formulation dominates across all datasets.

Taken together, these findings indicate that evidential design choices meaningfully affect calibration quality and selective performance in ways that are not reflected by accuracy alone. Model behavior depends strongly on dataset scale, class count, and domain characteristics, reinforcing the need for systematic calibration analysis rather than reliance on a single evidential configuration.

\subsection{Experiment 2: Ensemble-Based Dirichlet Estimation via Method of Moments}\label{sec:exp2}

This experiment evaluates a stable alternative to evidential training by constructing Dirichlet concentration parameters from repeated softmax predictions. As described in Section~\ref{sec:momalphas}, we first obtain per-input Dirichlet parameters via method of moments from ensembles of CE–trained classifiers, then optionally refine them using fixed-point maximum likelihood iteration to assess whether likelihood-based adjustment improves uncertainty estimation beyond the moment approximation.

\subsubsection{Experimental Setup for Ensemble-Based Dirichlet Estimation}

We use the same four datasets as in Section~\ref{sec:exp1}: CIFAR-10, CIFAR-100, DermaMNIST, and APTOS. For each, we train $M=50$ models for 20 epochs (versus 30 epochs for EDL configurations) using the same ResNet-18 backbone and optimization protocol. This shorter training schedule reflects the differing convergence behavior of the two approaches. Classifiers trained with cross-entropy loss typically converge rapidly, and training for fewer epochs reduces computational cost when fitting many ensemble members while also helping avoid the excessive confidence that can arise from prolonged CE training. In contrast, EDL models generally require longer training to stabilize evidence accumulation, which benefits from the extended 30-epoch schedule used in Section~\ref{sec:exp1}.

Ensembles are formed by incrementally aggregating the first $M \in \{5, 10, 20, 30, 50\}$ independently trained models. Because each model is trained with a different random seed, the ordering is arbitrary; increasing $M$ in this manner simply corresponds to incorporating additional independently trained models without any structured ordering. For each ensemble size $M$, we first construct per-input Dirichlet parameters using the method of moments estimator described in Algorithm~\ref{alg:mom-dirichlet}. We then use these moment-based parameters as initializations for fixed-point maximum likelihood estimation, running 20 fixed-point updates per input to obtain likelihood-refined parameters described in Algorithm~\ref{alg:mle-dirichlet}. We choose 20 iterations as a conservative, fixed budget that is sufficient for stable convergence behavior in our setting while avoiding additional tuning across datasets and ensemble sizes.

CIFAR-100 and DermaMNIST results are shown in the main text, with CIFAR-10 and APTOS in Appendix~\ref{sec:append_exp2}. For each $M$, we report accuracy, F1 score, and NLL, alongside confidence histograms and reliability diagrams, using the same analysis protocol as Section~\ref{sec:exp1}. Because each method produces a single Dirichlet fit per ensemble size, results are reported without run-level averaging.

\subsubsection{Results for Ensemble-Based Dirichlet Estimation}

\begin{figure}[t]
    \centering
    \includegraphics[width=\textwidth]{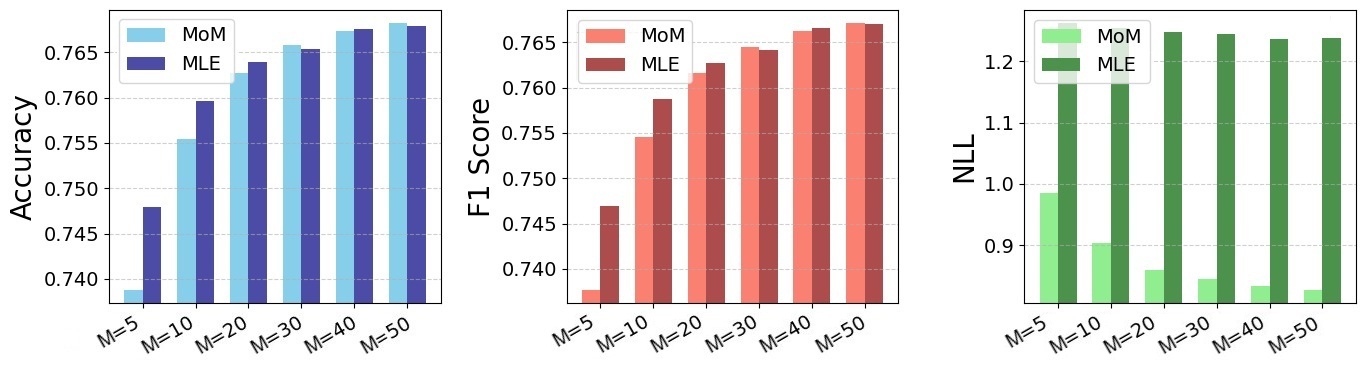}
    \caption[Predictive Performance Metrics for Ensemble-Based Dirichlet on CIFAR-100]{Predictive performance metrics (accuracy, macro F1, and negative log-likelihood) for the evaluated ensemble-based Dirichlet estimators on CIFAR-100.}
    \label{fig:C100_mom_metrics}
\end{figure}

\begin{figure}[!t]
    \centering
    \includegraphics[width=\textwidth]{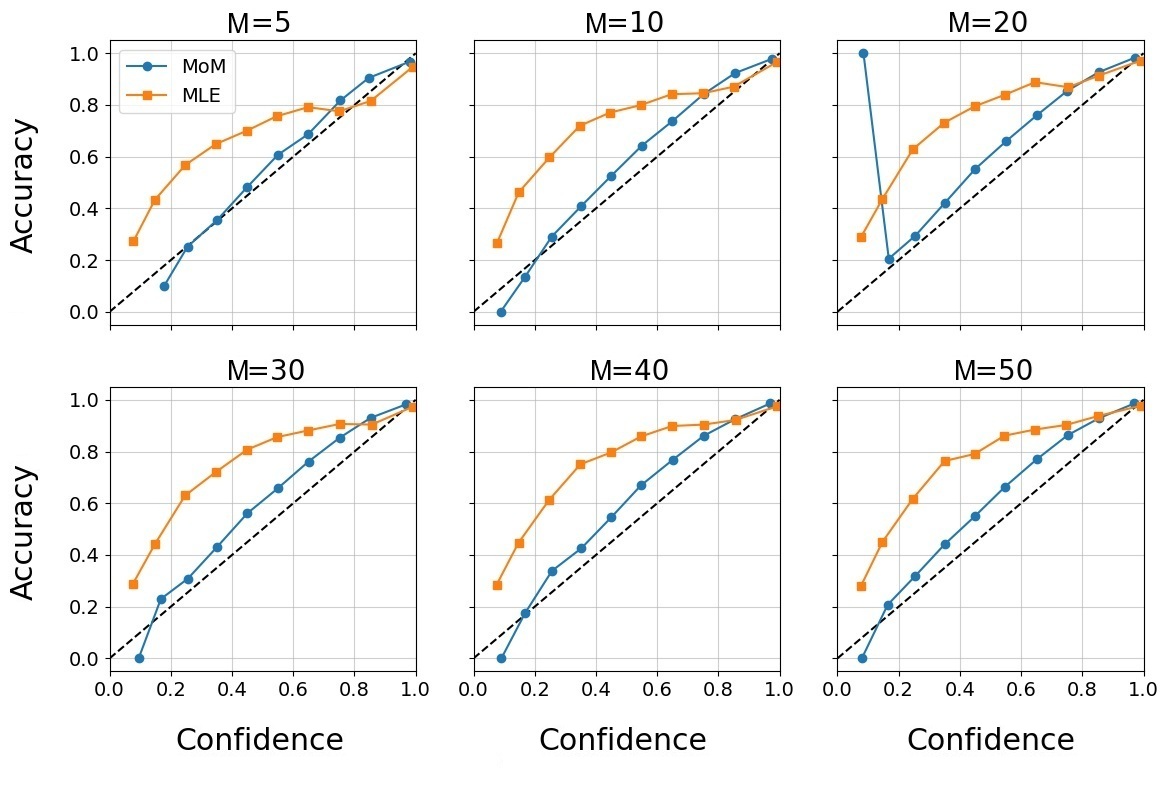}
    \caption[Reliability Diagrams for Ensemble-Based Dirichlet Estimators on CIFAR-100]{Reliability diagrams for ensemble-based Dirichlet estimators on CIFAR-100. The plots compare calibration behavior of method of moments and likelihood-refined Dirichlet parameter estimates across ensemble sizes $M$, where $M$ denotes the number of independently trained models aggregated in the ensemble. Each curve plots empirical accuracy against predicted confidence, where perfect calibration corresponds to the diagonal.}
    \label{fig:C100_MoM_rel}
\end{figure}

CIFAR-100 ensemble-based performance metrics are shown in Fig.~\ref{fig:C100_mom_metrics} and reliability diagrams in Fig.~\ref{fig:C100_MoM_rel}. For small ensemble sizes (five to ten models), the likelihood-refined estimates achieve higher accuracy and F1 scores than the ``coarser'' moment-based parameters. As the ensemble size increases, the two methods converge in predictive performance. In contrast, NLL remains consistently higher for the likelihood-refined estimates and shows limited improvement with larger ensemble sizes, whereas the moment-based estimates exhibit a steady decrease in NLL as aggregation increases. This difference is reflected in the reliability diagrams, where the likelihood-refined curves tend to lie well above the diagonal, indicating underconfident predictions relative to the moment-based estimates.

\begin{figure}
    \centering
    \includegraphics[width=\textwidth]{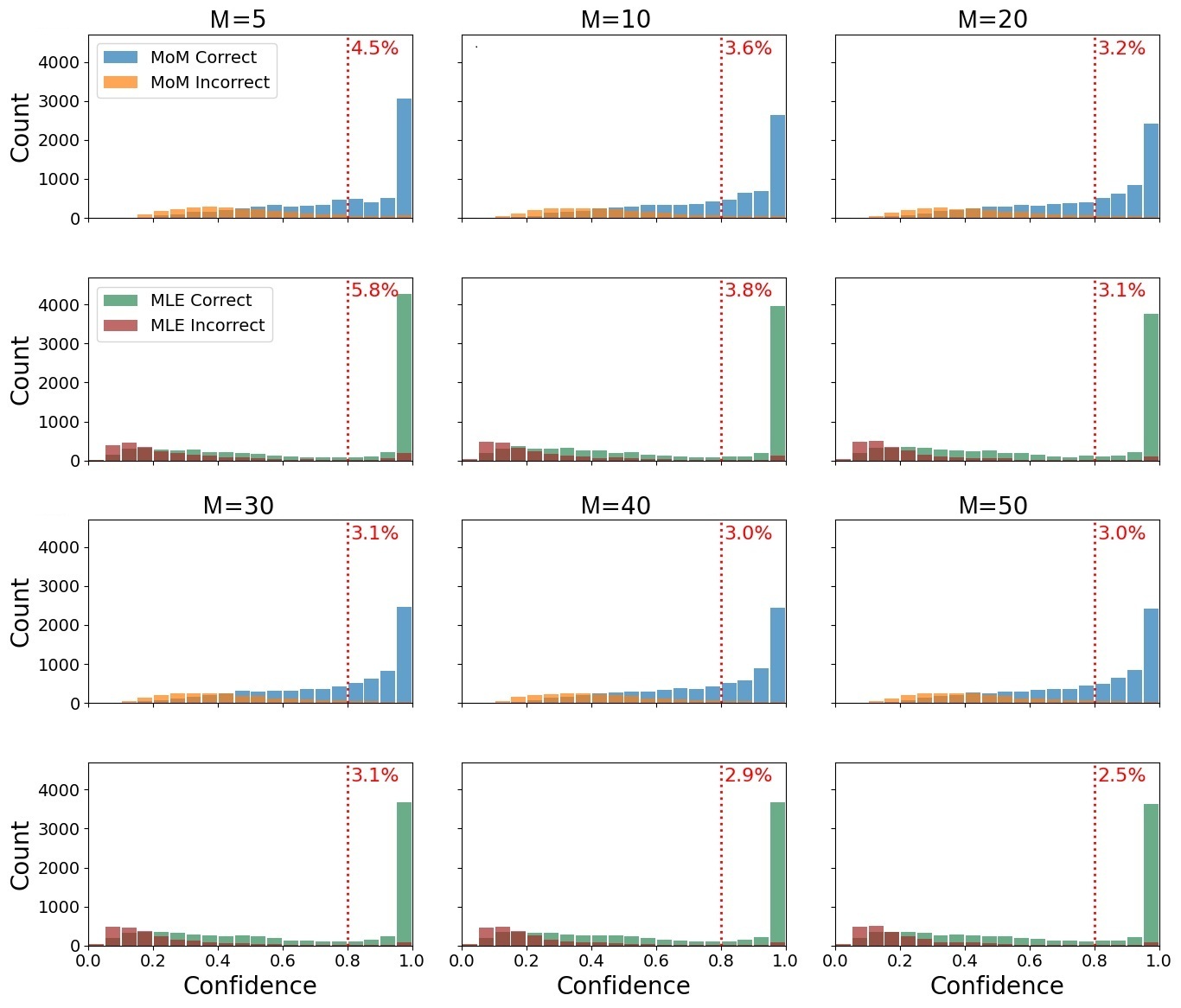}
    \caption[Confidence Calibration Diagnostics of Ensemble-Based Dirichlet on CIFAR-100]{CIFAR-100 confidence histograms for moment-based and likelihood-refined Dirichlet estimators across ensemble sizes. Histograms are conditioned on prediction correctness, with a vertical dashed line at the $0.8$ confidence threshold and the percentage of incorrect predictions above this threshold reported.}
    \label{fig:C100_mom_conf}
\end{figure}

The confidence histograms in Fig.~\ref{fig:C100_mom_conf} further illustrate this contrast. The moment-based estimates show strong separation between correct and incorrect predictions, with only a small portion of incorrect mass entering the high-confidence region. The likelihood-refined estimates, while producing fewer incorrect predictions above the $0.8$ threshold overall, exhibit a noticeable shift of correct predictions toward lower confidence and a concentration of incorrect mass in the extreme high-confidence tail. This redistribution of confidence aligns with the elevated NLL observed for the likelihood-refined parameters. 

\begin{figure}[t]
    \centering
    \includegraphics[width=\textwidth]{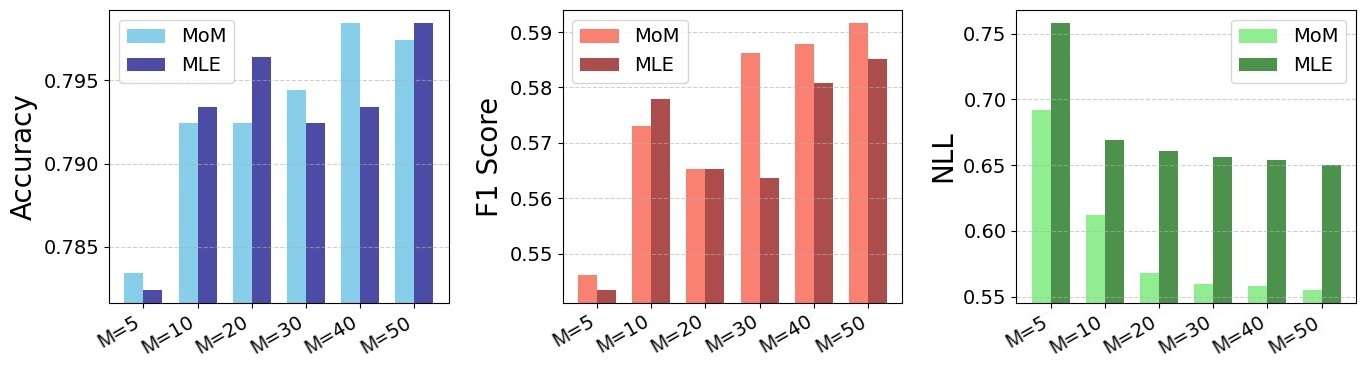}
    \caption[Predictive Performance Metrics for Ensemble-Based Dirichlet on DermaMNIST]{Predictive performance metrics (accuracy, macro F1, and negative log-likelihood) for the evaluated ensemble-based Dirichlet estimators on DermaMNIST.}
    \label{fig:DM_mom_metrics}
\end{figure}

Reviewing the DermaMNIST results, a less uniform pattern emerges, though several themes mirror those observed on CIFAR-100. In Fig.~\ref{fig:DM_mom_metrics}, the moment-based accuracy and F1 scores increase more steadily with ensemble size, whereas the likelihood-refined metrics fluctuate as additional models are aggregated. NLL decreases for both methods as ensemble size grows, but the moment-based estimates attain lower values earlier and maintain a slight advantage across most aggregation levels.

\begin{figure}[!t]
    \centering
    \includegraphics[width=\textwidth]{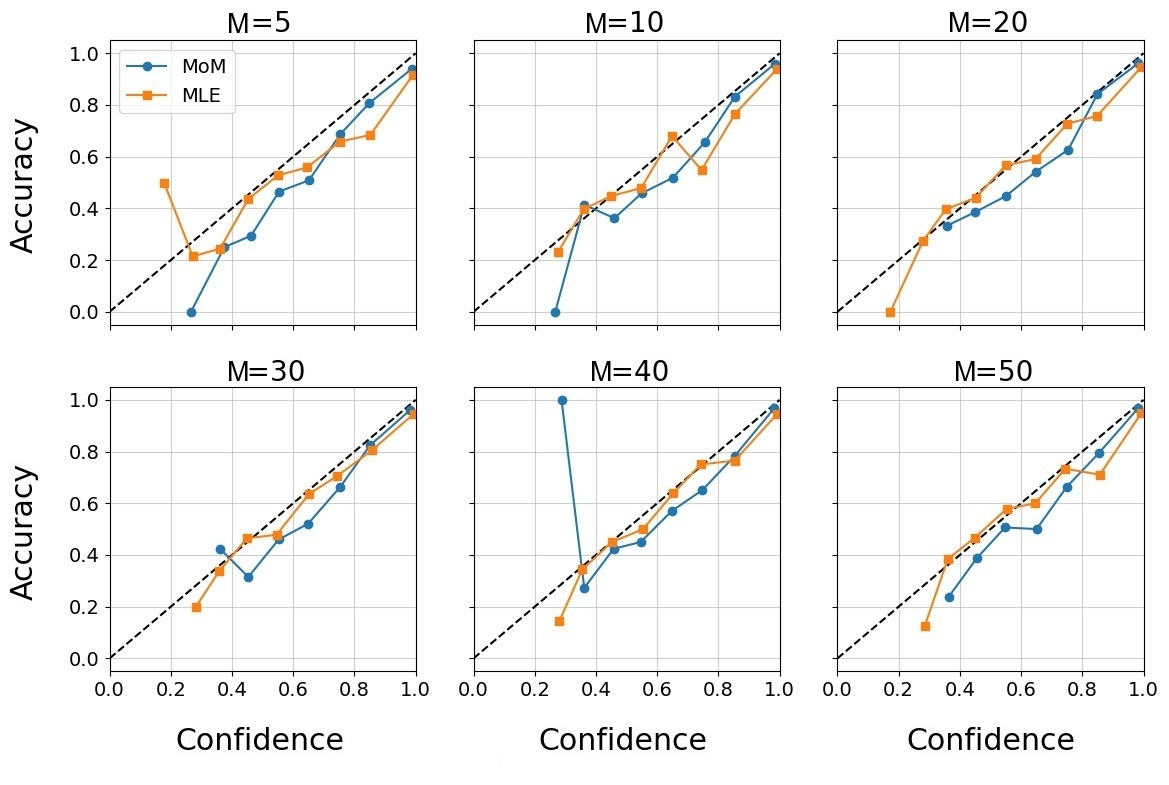}
    \caption[Reliability Diagrams for Ensemble-Based Dirichlet Estimators on DermaMNIST]{Reliability diagrams for ensemble-based Dirichlet estimators on DermaMNIST. The plots compare calibration behavior of method of moments and likelihood-refined Dirichlet parameter estimates across ensemble sizes $M$, where $M$ denotes the number of independently trained models aggregated in the ensemble. Each curve plots empirical accuracy against predicted confidence, where perfect calibration corresponds to the diagonal.}
    \label{fig:DM_MoM_rel}
\end{figure}

\begin{figure}
    \centering
    \includegraphics[width=\textwidth]{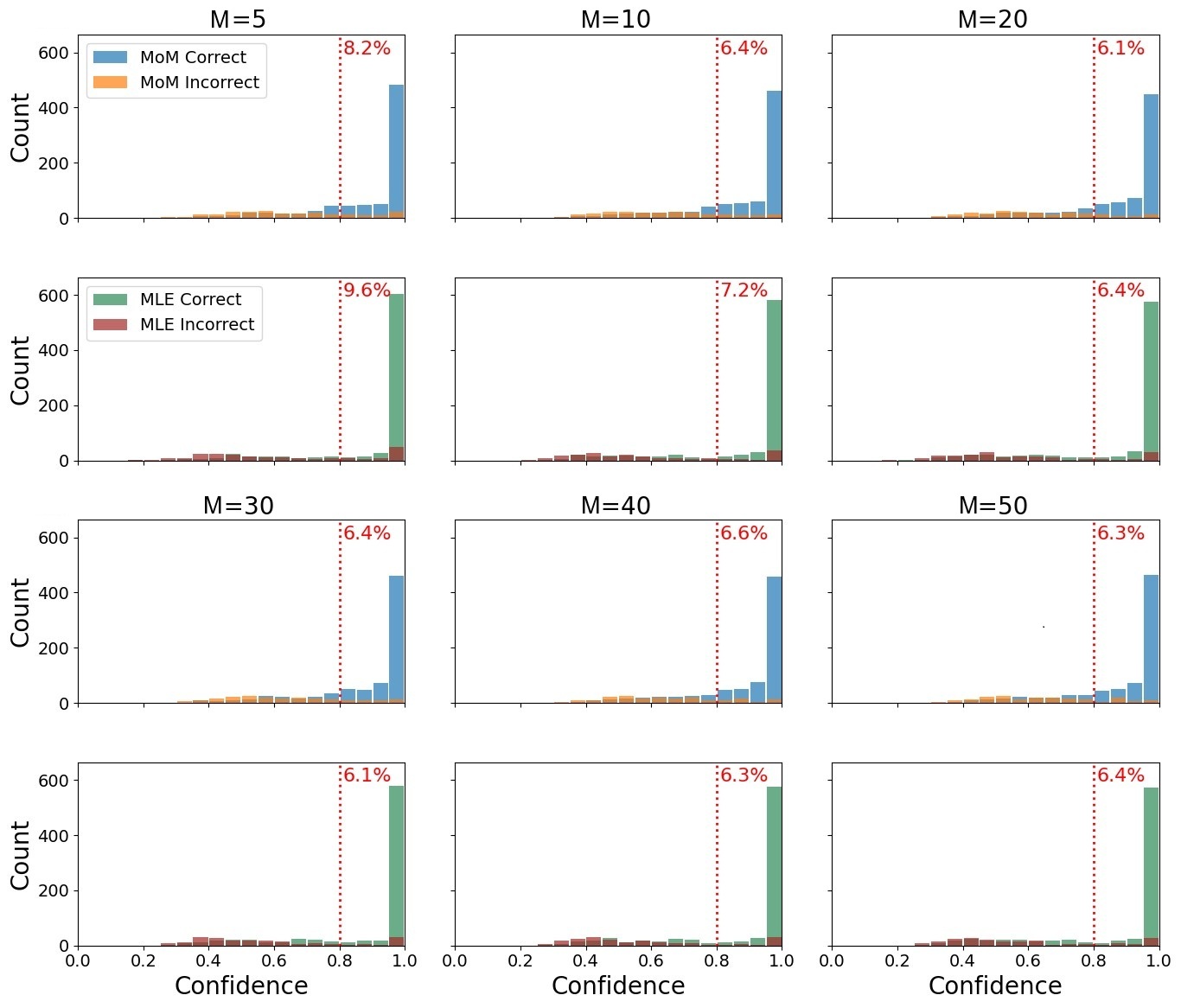}
    \caption[Confidence Calibration Diagnostics of Ensemble-Based Dirichlet on DermaMNIST]{DermaMNIST confidence histograms for moment-based and likelihood-refined Dirichlet estimators across ensemble sizes. Histograms are conditioned on prediction correctness, with a vertical dashed line at the $0.8$ confidence threshold and the percentage of incorrect predictions above this threshold reported.}
    \label{fig:DM_mom_conf}
\end{figure}

The reliability diagrams in Fig.~\ref{fig:DM_MoM_rel} show that the likelihood-refined estimates often track the diagonal reference line more closely, suggesting modest calibration improvements in certain confidence bins. However, the moment-based curves remain comparable, and in several bins the two methods intersect, indicating no consistent dominance of likelihood refinement as ensemble size increases.

Confidence histograms in Fig.~\ref{fig:DM_mom_conf} reinforce this conclusion. The moment-based estimates exhibit smoother separation between correct and incorrect predictions in the high-confidence region, without the pronounced peak of incorrect predictions observed under likelihood refinement. Although the overall percentage of incorrect predictions above the $0.8$ threshold remains similar between methods, the likelihood-refined estimates introduce sharper high-confidence concentration without a corresponding improvement in error control. This suggests limited practical benefit from the additional iterative computation in this setting.

Across all datasets (including those in the Appendices), likelihood refinement provides small accuracy and F1 gains only at very small ensemble sizes. As the ensemble grows, performance under moment-based and likelihood-refined estimates becomes effectively indistinguishable.  NLL shows a consistent pattern: likelihood refinement does not improve it and often produces higher values than the moment-based estimates. Reliability diagrams reveal no systematic calibration advantage, and confidence histograms show only minor, inconsistent reductions in high-confidence errors, often accompanied by undesirable redistribution of confidence mass.

Overall, the moment-based estimator produces calibration and selective classification performance comparable to or better than likelihood refinement across ensemble sizes. The additional fixed-point refinement increases computation without delivering consistent improvements in predictive metrics or confidence calibration behavior. We therefore proceed with the moment-based construction alone in subsequent experiments.

\subsection{Experiment 3: Variance-Based Selective Classification}\label{sec:exp3}

This experiment uses DermaMNIST as a testbed for selective classification with Dirichlet-based uncertainty. As a small-scale medical imaging dataset with limited training data, it provides a realistic setting for evaluating uncertainty-guided decision control.  All models share a common backbone and optimization scheme to isolate the impact of uncertainty formulation. Dirichlet total predictive variance is used as the abstention signal. Its effectiveness is evaluated through risk–coverage curves and overlaid histograms of total variance for correct and incorrect predictions, enabling analysis of separation and ordering behavior.

\begin{figure}[t]
    \centering
    \begin{minipage}{0.49\linewidth}
        \centering
        \includegraphics[width=\linewidth]{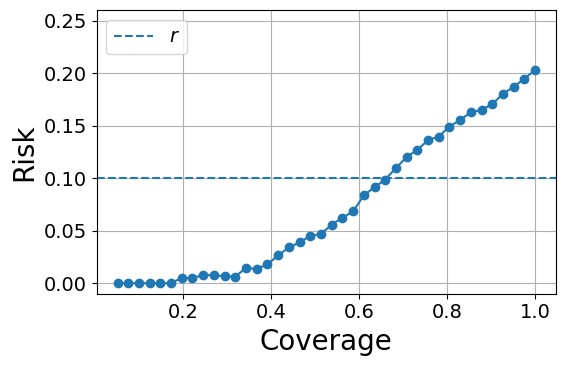}
    \end{minipage}
    \hfill
    \begin{minipage}{0.49\linewidth}
        \centering
        \includegraphics[width=\linewidth]{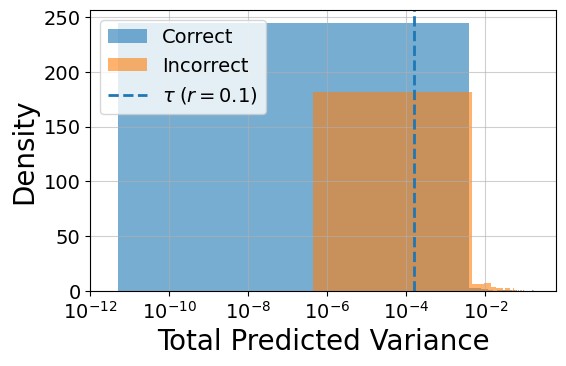}
    \end{minipage}
    \caption[Variance-Based Selective Classification via Ensemble-Based Dirichlet Construction]{Variance-based selective classification on DermaMNIST using ensemble-based Dirichlet parameters estimated via the method of moments. 
    \textbf{Left:} Risk–coverage curve with target risk $r = 0.1$ (dashed line) selected on a held-out calibration set. \textbf{Right:} Test set Dirichlet total predictive variance distributions, separated by correctness, with the corresponding abstention threshold $\tau$ indicated by a dashed line.}
    \label{fig:DM_MoM_sel}
\end{figure}

On the left side of Fig.~\ref{fig:DM_MoM_sel}, the method of moments parameter estimates achieve a retained-set accuracy of $89.9\%$ at a calibrated risk threshold of $r = 0.1$, retaining $70\%$ of the test set. This demonstrates a favorable trade-off between accuracy and coverage.  On the right side, the overlaid variance histogram shows meaningful separation between total predictive variance for correct and incorrect predictions. Correct predictions concentrate in the low-variance region, while incorrect predictions cluster toward higher variance. The threshold $\tau$, selected to satisfy the fixed risk target $r = 0.1$, lies near the transition between these distributions. Although some incorrect predictions remain above the threshold and some correct predictions are discarded, the separation supports total predictive variance as an effective abstention criterion.

To contextualize these results, we compare the ensemble-based Dirichlet construction with two evidential classifiers that exhibit contrasting calibration characteristics on DermaMNIST. As discussed in Section~\ref{sec:exp1}, the Exponential configuration achieved the most stable and favorable calibration metrics across runs, whereas the Digamma configuration demonstrated unstable calibration behavior despite comparable classification accuracy. We therefore apply the same variance-based selective classification procedure to these evidential variants for direct comparison.

\begin{figure}[!t]
    \centering
    \subfloat[Risk–Coverage: Digamma vs. Exponential]{
        \includegraphics[width=\textwidth]{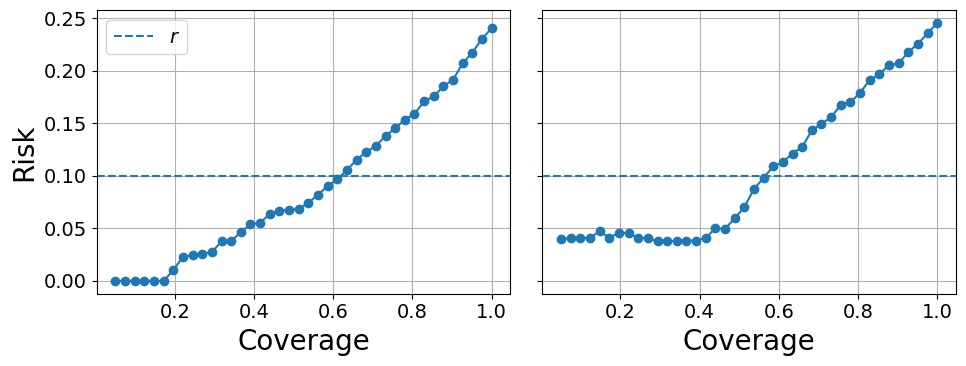}
        \label{fig:DM_EDL_riskcurve}
    }\\[1.5mm]

    \subfloat[Variance Distributions by Correctness]{
        \includegraphics[width=\textwidth]{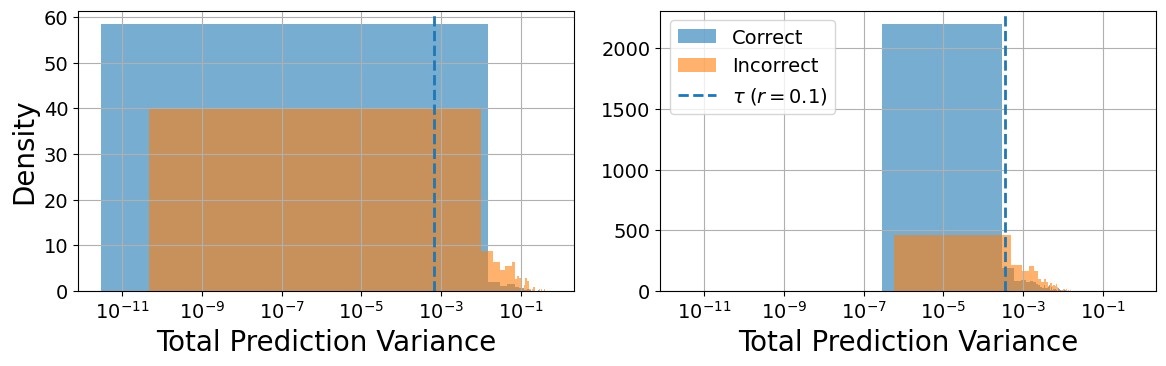}
        \label{fig:DM_EDL_varhist}
    }\\[1.5mm]

    \caption[Variance-Based Selective Classification via EDL]{Variance-based selective classification on DermaMNIST for evidential Dirichlet models trained with Digamma (left column) and Exponential (right column) losses.
    \textbf{(a)} Risk–coverage curves with target risk $r$ $=$ $0.1$ (dashed line) and thresholds selected on a held-out calibration set.
    \textbf{(b)} Test set Dirichlet total predictive variance distributions, separated by correctness, with the corresponding abstention threshold indicated.}
    \label{fig:DM_EDL_sel}
\end{figure}

Fig.~\ref{fig:DM_EDL_riskcurve} presents the risk–coverage trade-offs for the two evidential models. At the calibrated risk level $r = 0.1$, the Exponential model retains $56\%$ of the test set with a retained accuracy of $91.4\%$, while the Digamma model retains $63\%$ with $91.3\%$ accuracy. Both models therefore provide usable abstention signals and enable risk-controlled selective prediction at the chosen operating point.

The variance distributions in Fig.~\ref{fig:DM_EDL_varhist} offer additional context. Compared to the moment-based estimates in Fig.~\ref{fig:DM_MoM_sel}, the evidential variants exhibit greater overlap between correct and incorrect variance distributions, suggesting a weaker separation signal for thresholding. This difference is reflected in the numerical summary in Table~\ref{tab:dm_sel_summary}. Although variance-based thresholding remains viable for the evidential models, the separation between retained and rejected samples is less distinct.

From Table~\ref{tab:dm_sel_summary}, the method of moments parameters achieve the highest retained F1 score and the lowest retained NLL while retaining the largest fraction of the test set. Retained accuracy is slightly lower than the Exponential configuration but remains comparable. Overall, the ensemble-based construction provides competitive selective performance while maintaining stronger coverage and probabilistic quality.

All three approaches support variance-based selective classification at the calibrated risk level. However, the ensemble-based moment estimates yield a more separable variance signal and achieve similar or better retained metrics while preserving higher coverage. The evidential configurations remain competitive in retained accuracy, but their variance signals exhibit greater overlap, limiting thresholding efficiency.

\begin{table}[!t]
\centering
\caption[Comparison of Variance-Based Abstention Results on DermaMNIST]{DermaMNIST performance before and after variance-based abstention. Retained test metrics are computed after enforcing $r{=}0.1$ via a calibration-selected threshold. Best values per column are shown in bold. Here $N$ denotes the number of samples evaluated in each test subset, and $M$ denotes the number of independently trained models aggregated in the ensemble.}
\label{tab:dm_sel_summary}
\begin{tabular}{lcccccccc}
\hline
\textbf{Model} &
\multicolumn{4}{c}{\textbf{Full Test}} &
\multicolumn{4}{c}{\textbf{Retained Test}} \\
& $N$ & Acc & F1 & NLL & $N$ & Acc & F1 & NLL \\
\hline
Digamma
& 1002 & 0.7605 & 0.3963 & 1.7061
& 634  & 0.9132 & 0.4346 & 1.1109 \\

Exponential
& 1002 & 0.7515 & 0.4366 & 0.7215
& 559  & \textbf{0.9141} & 0.5204 & 0.3496 \\

MoM Alphas ($M{=}50$)
& 1002 & \textbf{0.7974} & \textbf{0.5916} & \textbf{0.5552}
& \textbf{701}  & 0.8987 & \textbf{0.6292} & \textbf{0.3124} \\
\hline
\end{tabular}
\end{table}

\section{Discussion}\label{sec:momdiscussion}

This work revealed a practical concern: models trained with cross-entropy loss can produce accurate point predictions, yet their softmax scores fluctuate in scale across training runs and offer no intrinsic uncertainty representation. Although evidential formulations aim to address this limitation by modeling class probabilities with Dirichlet distributions, we showed that their behavior depends strongly on implementation choices. Across datasets, loss selection, regularization, and evidence activation materially influence calibration quality and selective classification performance, even when accuracy remains similar.

In Experiment 1, we conducted a systematic empirical analysis of Dirichlet-based classifiers for uncertainty-aware decision making. The findings demonstrate that evidential configurations with nearly identical accuracy can diverge meaningfully in NLL, reliability behavior, and high-confidence error rates. No single formulation dominates across natural image, high-class-count, and medical datasets. Calibration performance depends on dataset scale and domain characteristics, reinforcing that uncertainty evaluation requires more than simple accuracy assessments.

In Experiment 2, we investigated a method of moments estimator for constructing Dirichlet distributions from ensembles of CE–trained models and evaluated an optional maximum likelihood refinement via fixed-point iteration. Likelihood refinement provides modest gains only for very small ensembles. As ensemble size increases, moment-based and likelihood-refined estimates become effectively indistinguishable in accuracy and F1 score. NLL does not improve under refinement and is often slightly higher. Reliability diagrams and confidence histograms reveal no consistent calibration advantage. Because the computational cost of refinement grows with ensemble size while performance differences diminish, the moment-based construction offers a more efficient and stable alternative for practical use.

In Experiment 3, we demonstrated that ensemble-derived Dirichlet variance serves as an effective ordering statistic for selective classification. Under matched risk constraints, the moment-based construction produces a variance signal with clearer separation between correct and incorrect predictions, enabling competitive or improved retained performance at comparable or higher coverage relative to evidential baselines. Evidential models remain viable, but their variance distributions exhibit greater overlap, reducing thresholding efficiency.

Taken together, these results show that uncertainty estimates derived from evidential training must be carefully analyzed, as their behavior depends strongly on loss design, regularization, and activation choices. When evidential configurations are well specified, they can yield competitive calibration and selective performance; however, when they are unstable or poorly configured, uncertainty quality degrades. The ensemble-based Dirichlet construction provides a stable alternative for generating concentration parameters without reliance on evidential loss design, while likelihood refinement offers limited additional benefit beyond the moment-based estimates in this setting.

\backmatter

\section*{Declarations}

\begin{itemize}
\item Disclaimer: The views expressed in this article are those of the author and do not reflect the official policy or position of the Department of the Air Force, Department of War, or the U.S. Government.
\item Funding: The authors received no specific funding for this work.
\item Conflict of interest/Competing interests: The authors declare that they have no competing interests.
\item Ethics approval and consent to participate: Not applicable. This study uses publicly available, de-identified datasets and does not involve human subjects research requiring institutional review board approval.
\item Consent for publication: All authors have read and approved the final manuscript and consent to its publication.
\item Data availability: All datasets used in this study are publicly available and properly cited in the manuscript. Access information is provided in the corresponding references.
\item Materials availability: No additional materials were used beyond the publicly available datasets and standard machine learning frameworks described in the manuscript.
\item Code availability: The code used to reproduce the experiments and figures in this study will be made publicly available in a GitHub repository upon publication.
\item Author contribution: Both authors conceived the study, designed and conducted the experiments, implemented the methodology, analyzed the results, and wrote the manuscript.
\end{itemize}

\begin{appendices}

\section{Loss Function Derivations}\label{sec:lossders}

The derivations of the loss functions vary in complexity but rely on standard identities of the Dirichlet distribution to obtain simplified forms. Let $\mathbf{p}$ denote a random probability vector distributed according to a Dirichlet distribution with concentration parameters $\boldsymbol{\alpha}$ (i.e., $\mathbf{p} \sim \mathrm{Dir}(\boldsymbol{\alpha})$). Throughout the following derivations, expectations and variances are taken with respect to this distribution and are denoted by $\mathbb{E}[\cdot]$ and $\mathbb{V}[\cdot]$, respectively.

We begin with the more straightforward derivation of the MSE loss, where
\begin{align}\label{eq:appmseloss}
    \mathcal{L}_{\mathrm{MSE}}
    &= \int \| \mathbf{y} - \mathbf{p} \|^2_2 \; \frac{1}{\mathrm{B}(\boldsymbol{\alpha})} \prod_{k=1}^K p_k^{\alpha_k - 1} \; d\boldsymbol{p} \notag \\
    &= \mathbb{E}\left[ \| \mathbf{y} - \mathbf{p} \|^2_2 \right] \notag \\
    &= \mathbb{E}\left[\sum_{k=1}^K \left( y_k^2 - 2y_k p_k +p_k^2\right)\right] \notag \\
    &= \sum_{k=1}^K \left(y_k^2 - 2y_k\mathbb{E}[p_k] +\mathbb{E}[p_k^2]\right) \notag \\
    &= \sum_{k=1}^K \left(y_k^2 - 2y_k\frac{\alpha_k}{\alpha_0} +\left(\mathbb{V}[p_k]+\mathbb{E}[p_k]^2\right)\right) \notag \\
    &= \sum_{k=1}^K \left(y_k^2 - 2y_k\frac{\alpha_k}{\alpha_0}  \frac{\alpha_k(\alpha_0-\alpha_k)}{\alpha_0^2(\alpha_0+1)} + \left(\frac{\alpha_k}{\alpha_0}\right)^2\right) \notag \\
    &= \sum_{k=1}^K \left[\left( y_k - \frac{\alpha_k}{\alpha_0} \right)^2 + \frac{\alpha_k(\alpha_0 - \alpha_k)}{\alpha_0^2(\alpha_0+1)}\right].
\end{align}

The derivation of the digamma loss requires a few more identities and steps.  First, we start with the expectation expression  
\begin{align}
    \mathcal{L}_{\mathrm{Digamma}} 
    &= \mathbb{E}
    \left[-\sum_{k=1}^K y_k \log p_k\right] \notag \\
    &= -\sum_{k=1}^K y_k \; \mathbb{E} [\log p_k] \notag \\
    &= -\sum_{k=1}^K y_k \frac{1}{\mathrm{B}(\boldsymbol{\alpha})} \int \log p_k \prod_{i=1}^K p_i^{\alpha_i-1} d\boldsymbol{p}.
\end{align}

To simplify the evaluation of the remaining integral, we differentiate the multivariate Beta function with respect to $\alpha_k$ to produce a useful identity:  
\begin{align}
    \mathrm{B}(\boldsymbol{\alpha}) &=\int \prod_{i=1}^K p_i^{\alpha_i-1} d\boldsymbol{p} \notag \\
    \frac{\partial\mathrm{B}(\boldsymbol{\alpha}) }{\partial{\alpha_k}}  &= \int \frac{\partial}{\partial{\alpha_k}}\left(\prod_{i=1}^K p_i^{\alpha_i-1} \right)d\boldsymbol{p} \notag \\
    &= \int \prod_{i=1}^K p_i^{\alpha_i-1} \log{p_k} \; d\boldsymbol{p}
\end{align}

Thus, substituting this identity into our original integral in Eq.~\eqref{eq:appdigammaloss} and applying the definition of the multivariate Beta function (see \eqref{eq:mvbfunc}) along with some calculus and algebraic identities, we get
\begin{align}
    \mathcal{L}_{\mathrm{Digamma}} &= -\sum_{k=1}^K y_k  \frac{1}{\mathrm{B}(\boldsymbol{\alpha})}\frac{\partial\mathrm{B}(\boldsymbol{\alpha}) }{\partial{\alpha_k}}  \notag \\
    &= -\sum_{k=1}^K y_k \frac{\partial}{\partial{\alpha_k}}\log{\mathrm{B}(\boldsymbol{\alpha})}  \notag \\
    &= -\sum_{k=1}^K y_k \left(\frac{\partial}{\partial{\alpha_k}}\sum_{i=1}^K{\log{\Gamma(\alpha_i)}}-\frac{\partial}{\partial{\alpha_k}}\log{\Gamma(\alpha_0)}  \right) \notag \\
    &= -\sum_{k=1}^K y_k \left(\frac{\partial{}}{\partial{\alpha_k}}{\log{\Gamma(\alpha_k)}}-\frac{\partial{\alpha_0}}{\partial{\alpha_k}} \cdot \frac{\partial}{\partial{\alpha_0}} \log{\Gamma(\alpha_0)} \right)
\end{align}

We note that $\partial{\alpha_0}/\partial{\alpha_k}=1$ and the definition of the digamma function is
\begin{equation}
    \psi(\alpha_k) := \frac{\partial}{\partial{\alpha_k}}\log \Gamma(\alpha_k),
\end{equation}
which produces the final simplified form referred to as the digamma loss, 
\begin{align}\label{eq:appdigammaloss}
    \mathcal{L}_{\mathrm{Digamma}} &= \sum_{k=1}^K y_k \left(\psi{(\alpha_0)} - \psi{(\alpha_k)}\right).
\end{align}

\section{Experiment 1 Additional Figures}\label{sec:append_exp1}

\begin{figure}[t]
    \centering
    \includegraphics[width=\textwidth]{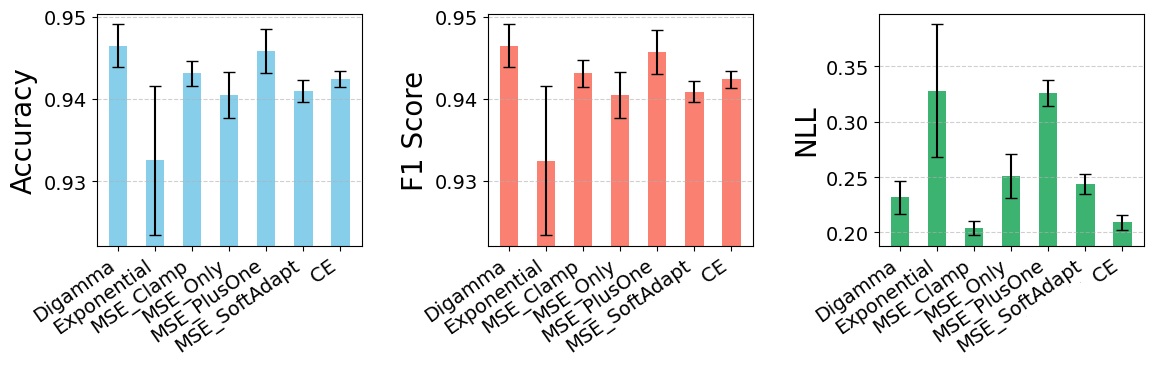}
    \caption[EDL Predictive Performance Metrics for CIFAR-10]{Predictive performance metrics (accuracy, macro F1, and negative log-likelihood) for the evaluated EDL configurations on CIFAR-10.}
    \label{fig:c10_EDL_metrics}
\end{figure}

\begin{figure}[t]
    \centering
    \begin{subfigure}{0.5\linewidth}
        \includegraphics[width=\textwidth]{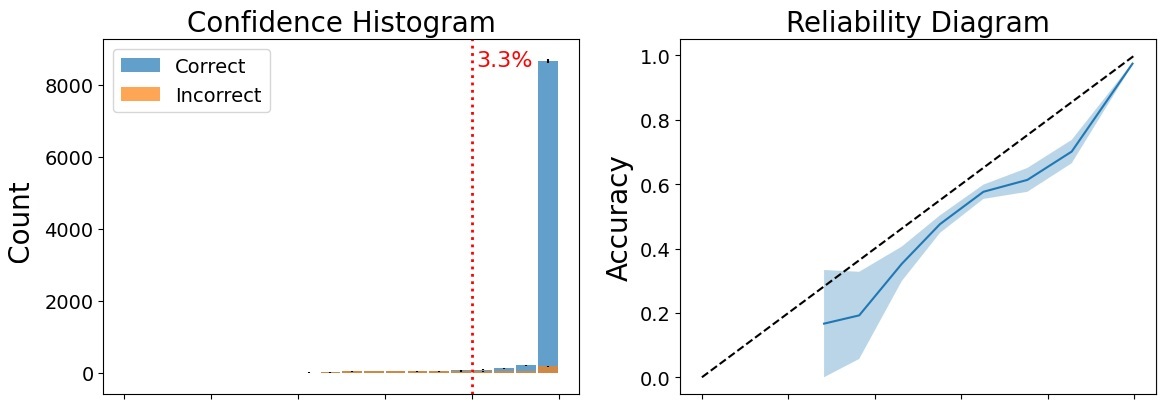}
        \caption{MSE Only}
        \label{fig:c10_mse_only}
    \end{subfigure}\hfill
    \begin{subfigure}{0.5\linewidth}
        \includegraphics[width=\textwidth]{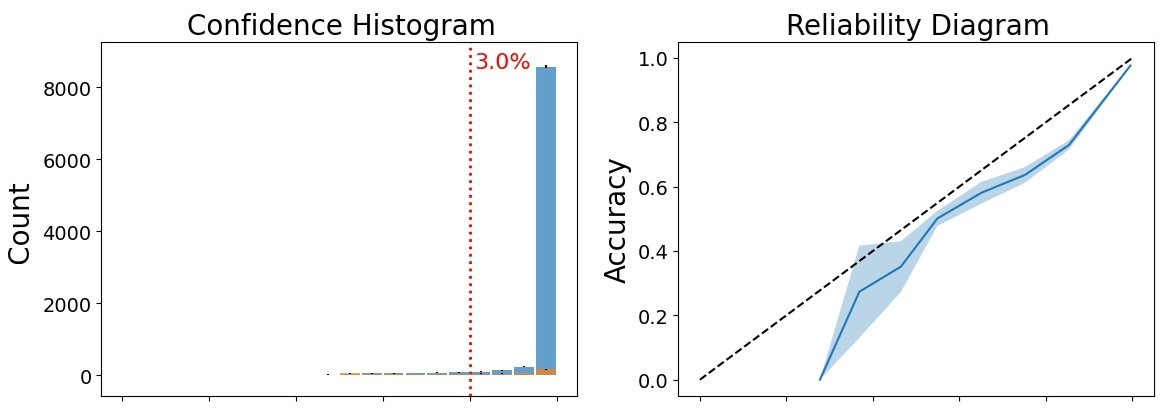}
        \caption{MSE SoftAdapt}
        \label{fig:c10_mse_softadapt}
    \end{subfigure}
    
    \begin{subfigure}{0.5\linewidth}
        \includegraphics[width=\textwidth]{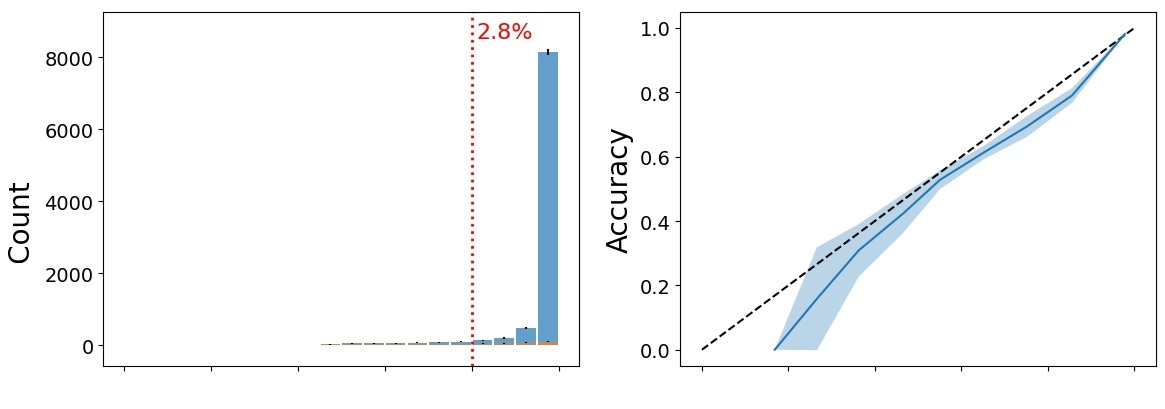}
        \caption{MSE Clamp}
        \label{fig:c10_mse_clamp}
    \end{subfigure}\hfill
    \begin{subfigure}{0.5\linewidth}
        \includegraphics[width=\textwidth]{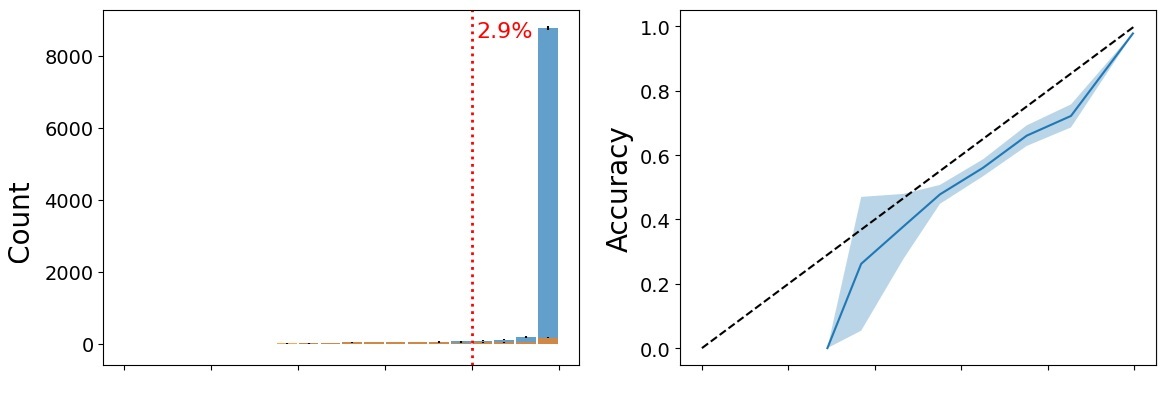}
        \caption{Digamma}
        \label{fig:c10_digamma}
    \end{subfigure}
    
    \begin{subfigure}{0.5\linewidth}
        \includegraphics[width=\textwidth]{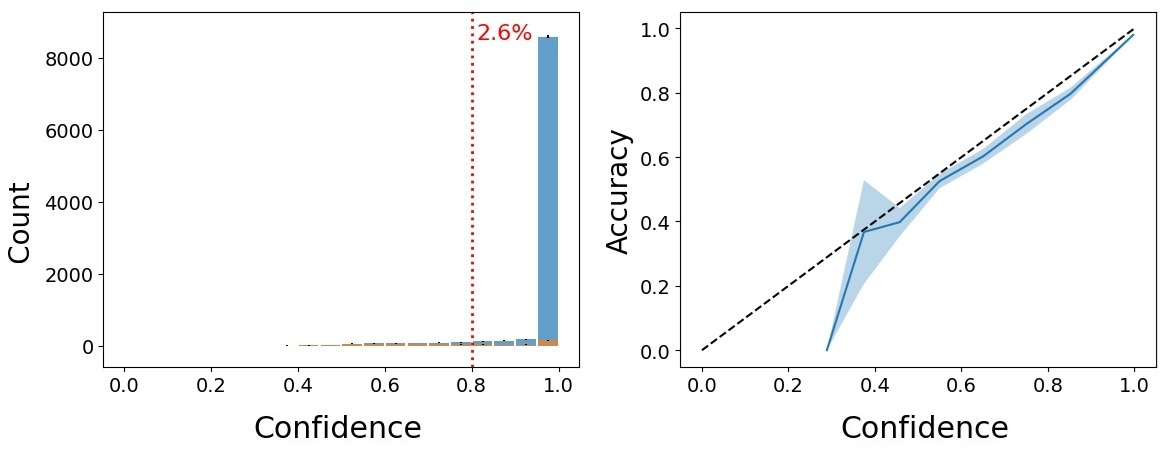}
        \caption{MSE PlusOne}
        \label{fig:c10_mse_plusone}
    \end{subfigure}\hfill
    \begin{subfigure}{0.5\linewidth}
        \includegraphics[width=\textwidth]{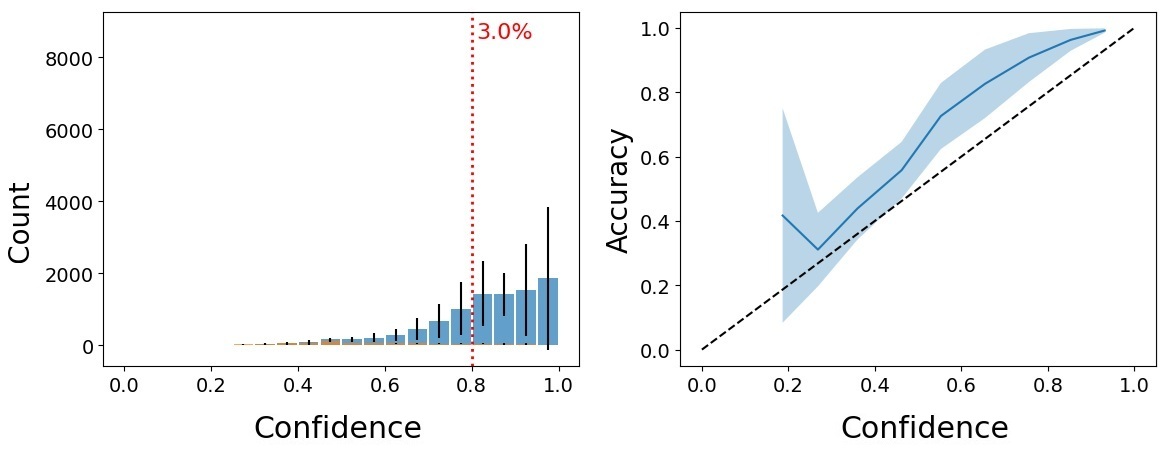}
        \caption{Exponential}
        \label{fig:c10_exponential}
    \end{subfigure}
    
    \caption[CIFAR-10 Confidence Calibration Diagnostics]{Confidence diagnostics for Dirichlet classifiers on CIFAR-10. Each subfigure corresponds to a different training configuration where the left panel shows the confidence histogram and the right panel shows the reliability diagram. Confidence histograms are computed from the mean predictions across runs, and the red annotation reports the percentage of incorrect predictions with confidence greater than $0.8$. Reliability diagrams summarize calibration behavior across runs, where shaded bands denote one standard deviation and the dashed line indicates perfect calibration.}
    \label{fig:c10_EDL_sum}
\end{figure}

\begin{figure}[t]
    \centering
    \includegraphics[width=\textwidth]{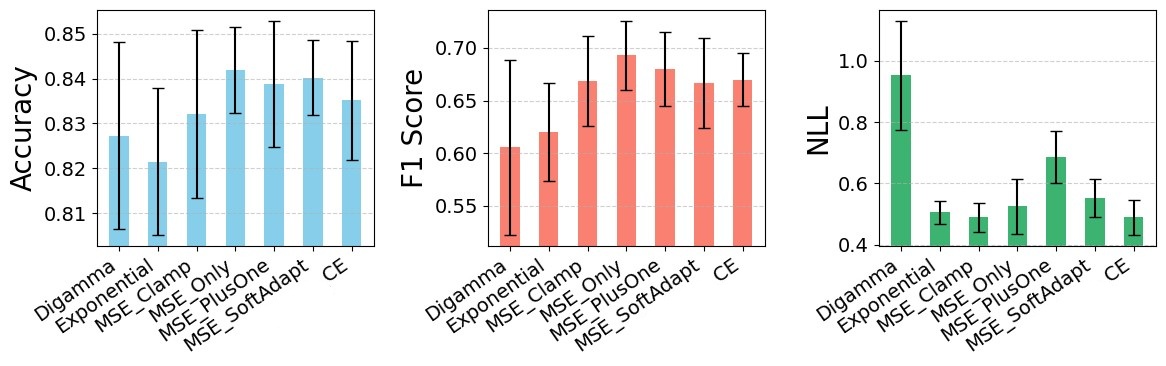}
    \caption[EDL Predictive Performance Metrics for APTOS]{Predictive performance metrics (accuracy, macro F1, and negative log-likelihood) for the evaluated EDL configurations on APTOS.}
    \label{fig:APT_EDL_metrics}
\end{figure}

\begin{figure}[t]
    \centering
    \begin{subfigure}{0.5\linewidth}
        \includegraphics[width=\textwidth]{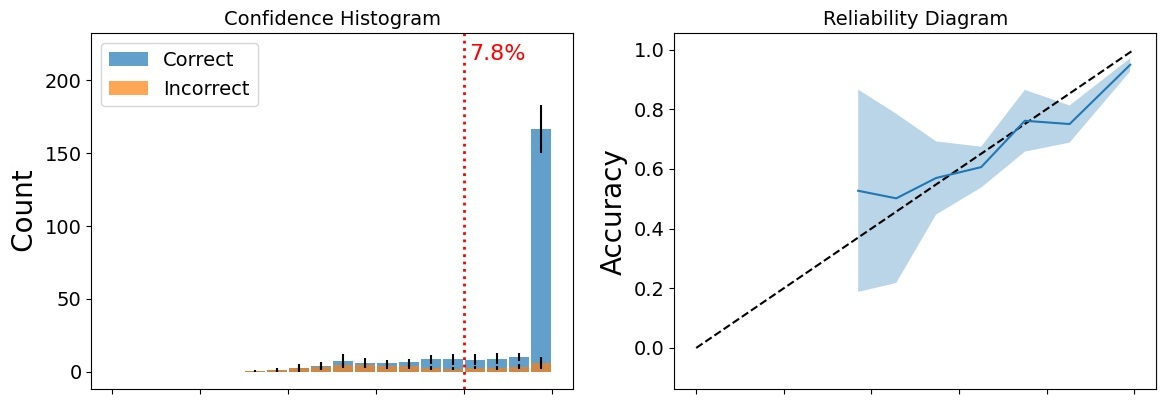}
        \caption{MSE Only}
        \label{fig:apt_mse_only}
    \end{subfigure}\hfill
    \begin{subfigure}{0.5\linewidth}
        \includegraphics[width=\textwidth]{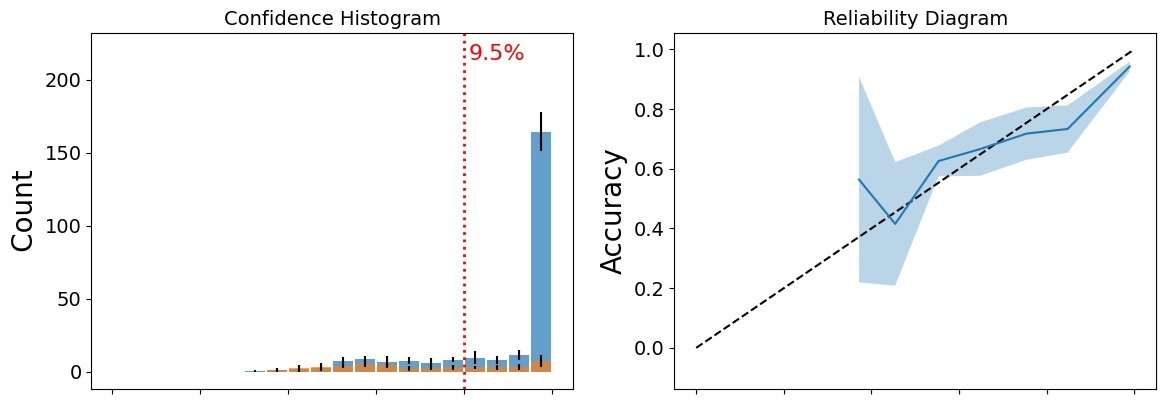}
        \caption{MSE SoftAdapt}
        \label{fig:apt_mse_softadapt}
    \end{subfigure}
    
    \begin{subfigure}{0.5\linewidth}
        \includegraphics[width=\textwidth]{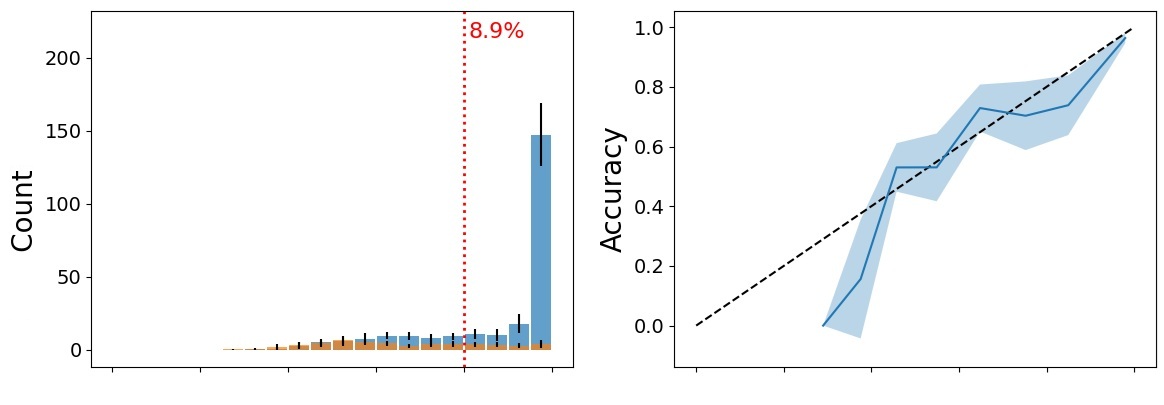}
        \caption{MSE Clamp}
        \label{fig:apt_mse_clamp}
    \end{subfigure}\hfill
    \begin{subfigure}{0.5\linewidth}
        \includegraphics[width=\textwidth]{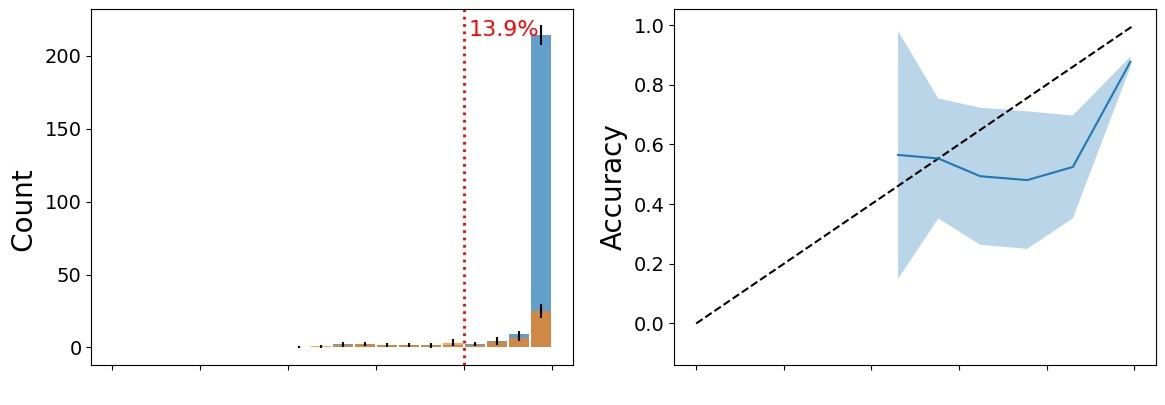}
        \caption{Digamma}
        \label{fig:apt_digamma}
    \end{subfigure}
    
    \begin{subfigure}{0.5\linewidth}
        \includegraphics[width=\textwidth]{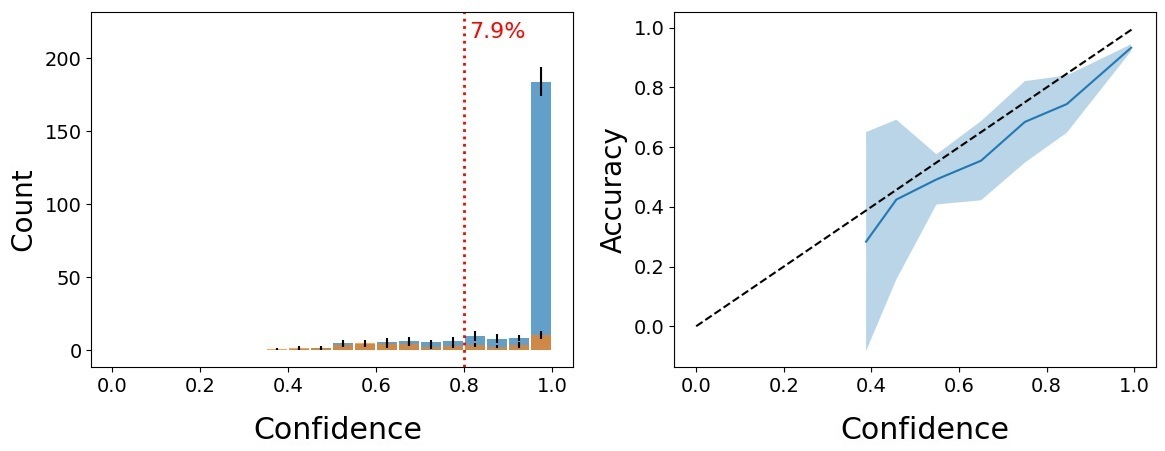}
        \caption{MSE PlusOne}
        \label{fig:apt_mse_plusone}
    \end{subfigure}\hfill
    \begin{subfigure}{0.5\linewidth}
        \includegraphics[width=\textwidth]{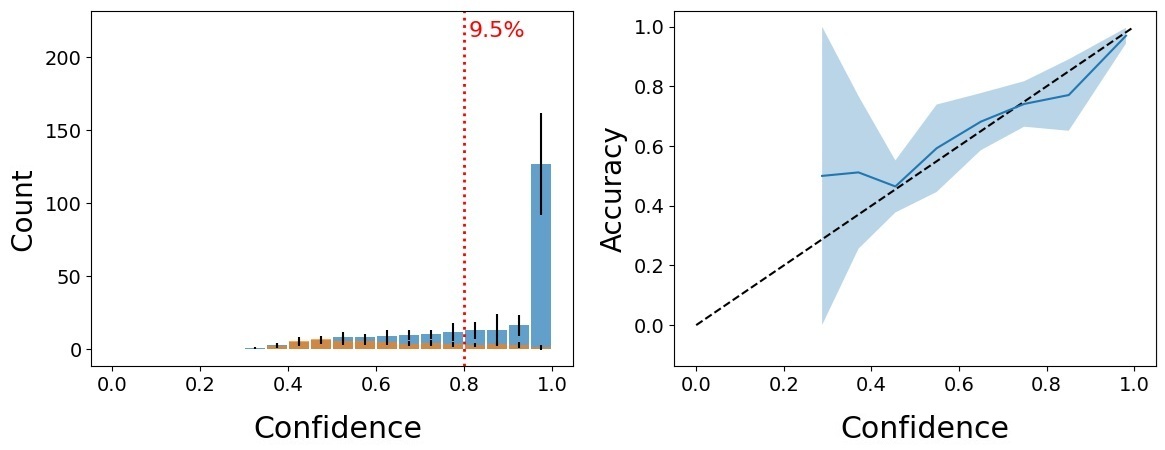}
        \caption{Exponential}
        \label{fig:apt_exponential}
    \end{subfigure}
    
    \caption[APTOS Confidence Calibration Diagnostics]{Confidence diagnostics for Dirichlet classifiers on APTOS. Each subfigure corresponds to a different training configuration where the left panel shows the confidence histogram and the right panel shows the reliability diagram. Confidence histograms are computed from the mean predictions across runs, and the red annotation reports the percentage of incorrect predictions with confidence greater than $0.8$. Reliability diagrams summarize calibration behavior across runs, where shaded bands denote one standard deviation and the dashed line indicates perfect calibration.}
    \label{fig:APT_EDL_sum}
\end{figure}

To supplement the primary results in Section~\ref{sec:exp1}, we include model metrics and calibration diagnostics for CIFAR-10 and APTOS in Figs.~\ref{fig:c10_EDL_metrics}\textendash\ref{fig:c10_EDL_sum} and~\ref{fig:APT_EDL_metrics}\textendash\ref{fig:APT_EDL_sum}, respectively. CIFAR-10 serves as a clean, low-class-count benchmark, while APTOS represents a more challenging medical imaging dataset characterized by class imbalance and limited training data.

On CIFAR-10, accuracy and F1 scores in Fig.~\ref{fig:c10_EDL_metrics} remain stable across configurations, with the exception of the Exponential variant. Negative log-likelihood exhibits greater variability, with Exponential producing the highest and most unstable values. In Fig.~\ref{fig:c10_EDL_sum}, MSE with clamping and MSE with $\delta=1$ display the most consistent reliability curves and the lowest proportion of incorrect predictions above the $0.8$ confidence threshold. Although Exponential deviates noticeably in the reliability diagrams and exhibits substantial variability, it does not produce the highest percentage of incorrect high-confidence predictions.

The APTOS results in Fig.~\ref{fig:APT_EDL_metrics} highlight the difficulty of this dataset. Accuracy and F1 scores show large standard deviations across runs, though mean performance remains similar across configurations. For negative log-likelihood, Digamma stands apart, producing both the highest values and the greatest instability. Calibration diagnostics in Fig.~\ref{fig:APT_EDL_sum} further reinforce this behavior: Digamma yields the largest proportion of incorrect high-confidence predictions and demonstrates pronounced overconfidence variability across independent training runs.

\clearpage

\section{Experiment 2 Additional Figures}\label{sec:append_exp2}

\begin{figure}[t]
    \centering
    \includegraphics[width=\textwidth]{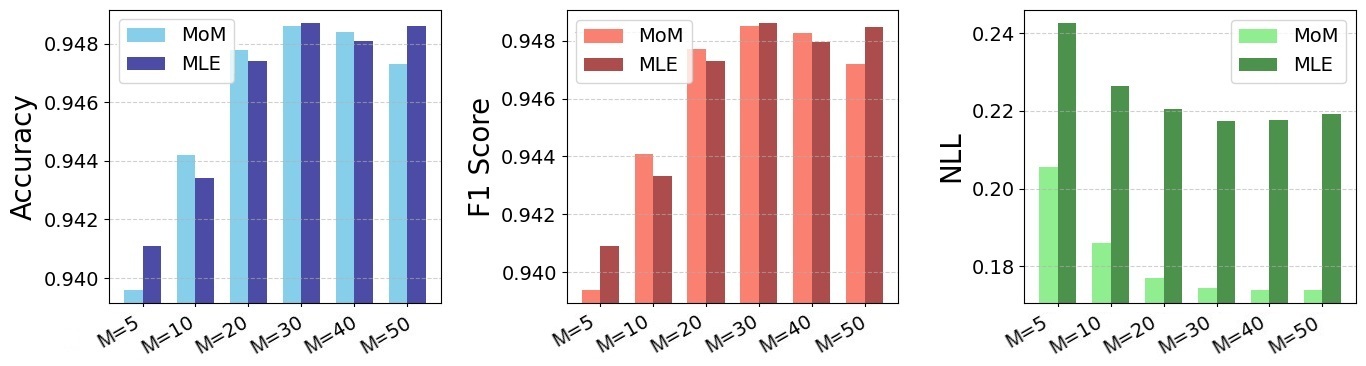}
    \caption[Ensemble-Based Dirichlet Predictive Performance Metrics for CIFAR-10]{Predictive performance metrics (accuracy, macro F1, and negative log-likelihood) for the evaluated ensemble-based Dirichlet estimators on CIFAR-10.}
    \label{fig:c10_mom_metrics}
\end{figure}

\begin{figure}[!t]
    \centering
    \includegraphics[width=\textwidth]{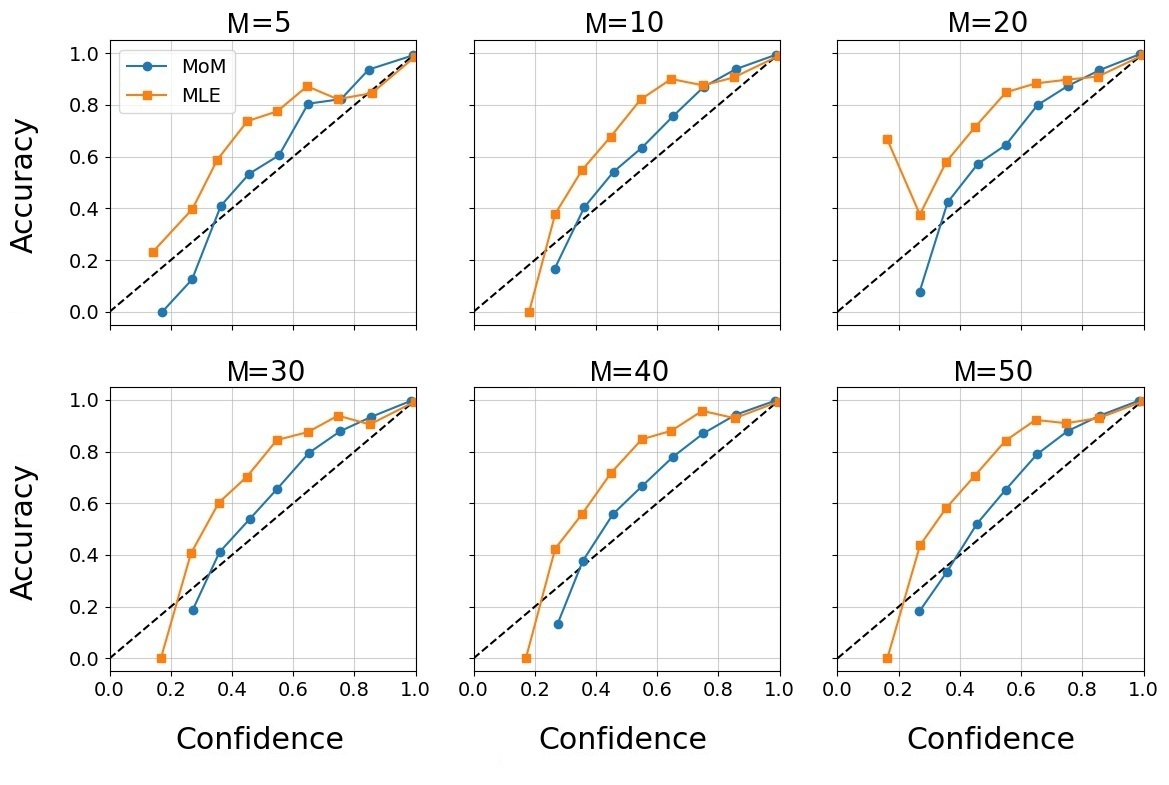}
    \caption[Reliability Diagrams for Ensemble-Based Dirichlet Estimators on CIFAR-10]{Reliability diagrams for ensemble-based Dirichlet estimators on CIFAR-10. The plots compare calibration behavior of method of moments and likelihood-refined Dirichlet parameter estimates across ensemble sizes $M$, where $M$ denotes the number of independently trained models aggregated in the ensemble. Each curve plots empirical accuracy against predicted confidence, where perfect calibration corresponds to the diagonal.}
    \label{fig:C10_MoM_rel}
\end{figure}

To supplement the primary results in Section~\ref{sec:exp2}, we include full calibration diagnostics for CIFAR-10 and APTOS.  For CIFAR-10 (Fig.~\ref{fig:c10_mom_metrics}), accuracy and F1 scores improve steadily with increasing ensemble size for both moment-based and likelihood-refined estimates, with results becoming nearly indistinguishable at larger aggregation levels. NLL follows the same trend observed in the primary experiments: likelihood-refined estimates remain consistently higher than moment-based values, though both decrease as ensemble size increases. The reliability diagrams in Fig.~\ref{fig:C10_MoM_rel} mirror the CIFAR-100 behavior, with likelihood-refined estimates exhibiting greater deviation from the diagonal reference line and a more underconfident profile relative to the moment-based estimates.

\begin{figure}
    \centering
    \includegraphics[width=\textwidth]{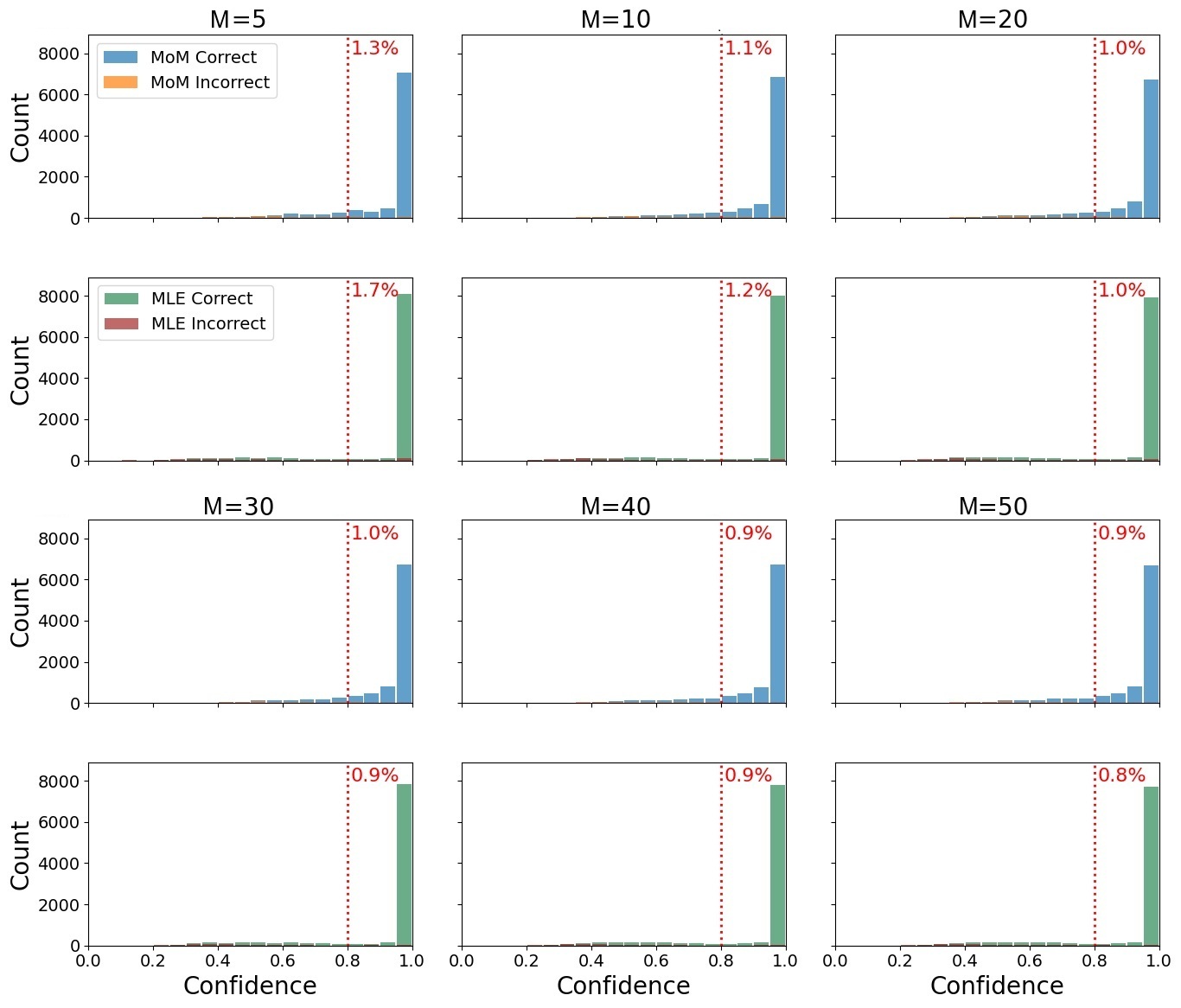}
    \caption{CIFAR-10 confidence histograms for moment-based and likelihood-refined Dirichlet estimators across ensemble sizes. Histograms are conditioned on prediction correctness, with a vertical dashed line at the $0.8$ confidence threshold and the percentage of incorrect predictions above this threshold reported.}
    \label{fig:C10_mom_conf}
\end{figure}

Confidence histograms in Fig.~\ref{fig:C10_mom_conf} show a modest reduction in incorrect high-confidence predictions under likelihood refinement. However, this comes with a pronounced concentration of mass in the top $5\%$ confidence region and a relative flattening of mid-range confidence values compared to the smoother distribution produced by the moment-based estimates.

\begin{figure}[t]
    \centering
    \includegraphics[width=\textwidth]{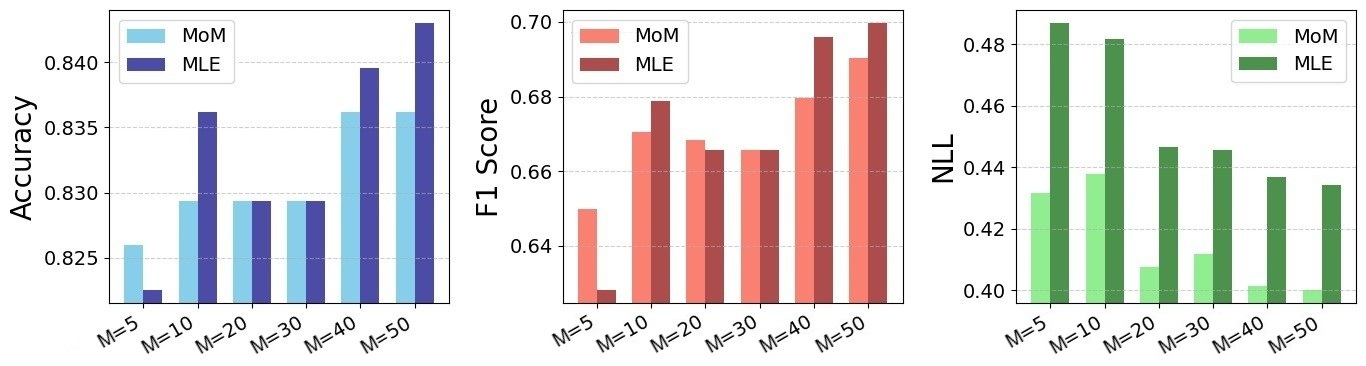}
    \caption[Ensemble-Based Dirichlet Predictive Performance Metrics for APTOS]{Predictive performance metrics (accuracy, macro F1, and negative log-likelihood) for the evaluated ensemble-based Dirichlet estimators on APTOS.}
    \label{fig:APT_mom_metrics}
\end{figure}

\begin{figure}[!t]
    \centering
    \includegraphics[width=\textwidth]{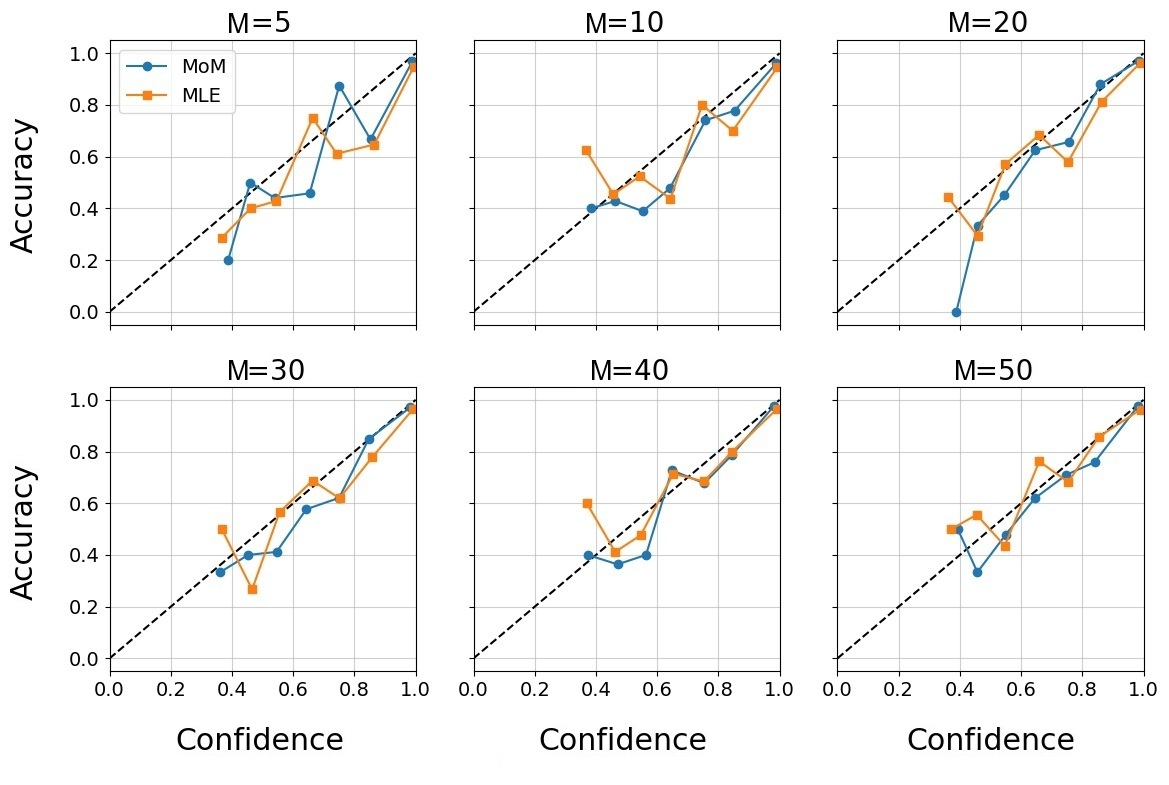}
    \caption[Reliability Diagrams for Ensemble-Based Dirichlet Estimators on APTOS]{Reliability diagrams for ensemble-based Dirichlet estimators on APTOS. The plots compare calibration behavior of method of moments and likelihood-refined Dirichlet parameter estimates across ensemble sizes $M$, where $M$ denotes the number of independently trained models aggregated in the ensemble. Each curve plots empirical accuracy against predicted confidence, where perfect calibration corresponds to the diagonal.}
    \label{fig:APT_MoM_rel}
\end{figure}

Turning to the APTOS metrics and reliability diagrams (Figs.~\ref{fig:APT_mom_metrics} and~\ref{fig:APT_MoM_rel}), both moment-based and likelihood-refined estimates exhibit instability at small ensemble sizes, with noticeable oscillations in accuracy, F1, and NLL. Although NLL decreases as the ensemble grows, the likelihood-refined estimates consistently remain higher than the moment-based values. The reliability diagrams show no consistent advantage for either method; both fluctuate around the diagonal reference line, with each approach closer to perfect calibration in different confidence bins.

The confidence histograms further reinforce this mixed behavior. Likelihood refinement yields a slight reduction in incorrect high-confidence predictions in some ensemble sizes, but the improvement is inconsistent and does not track with gains in predictive accuracy. Across ensemble sizes, both methods display a flattening of lower-confidence mass and intermittent spikes of correct and incorrect predictions in the higher-confidence region, suggesting limited practical benefit from the additional refinement step.

\begin{figure}
    \centering
    \includegraphics[width=\textwidth]{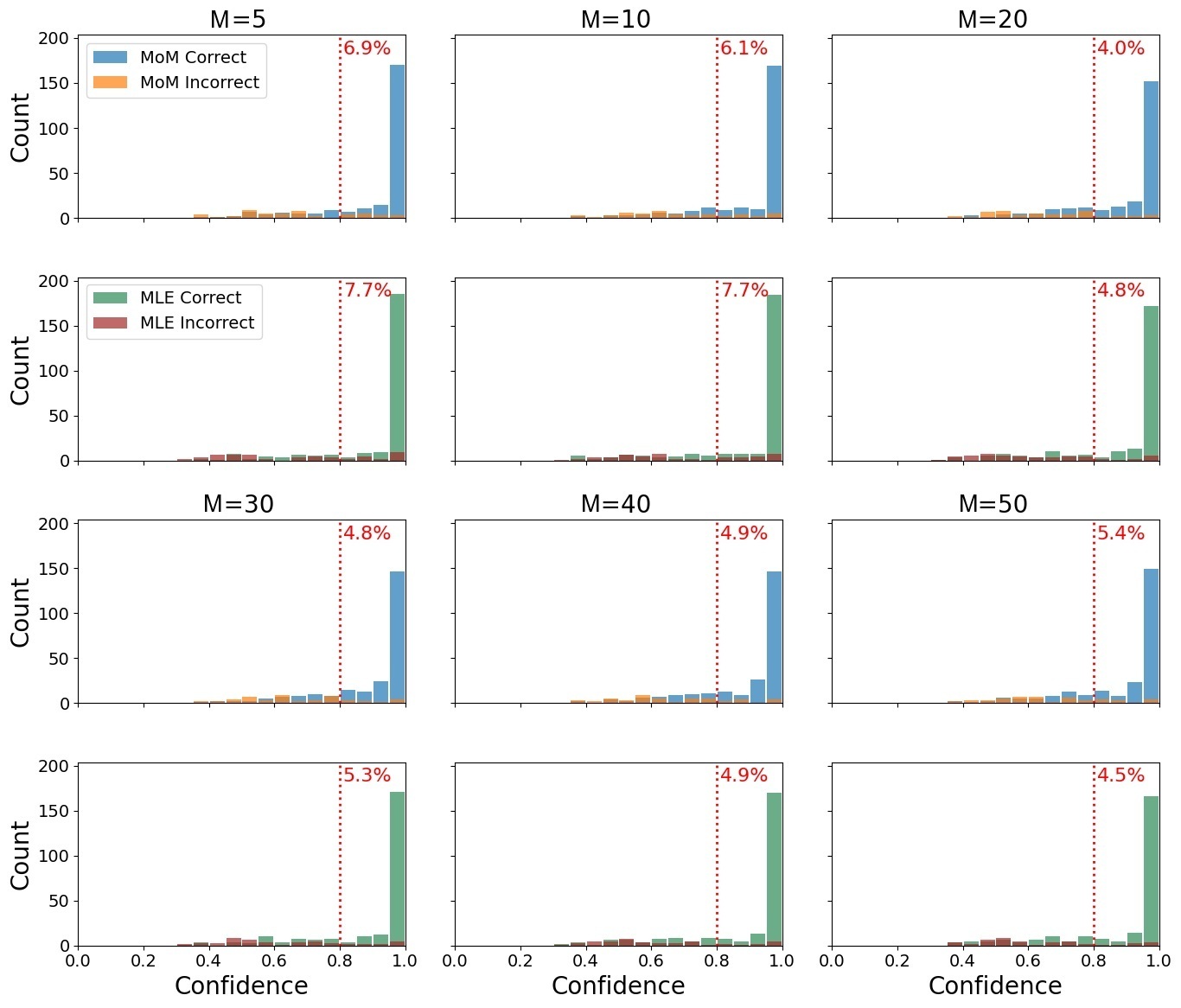}
    \caption{APTOS confidence histograms for moment-based and likelihood-refined Dirichlet estimators across ensemble sizes. Histograms are conditioned on prediction correctness, with a vertical dashed line at the $0.8$ confidence threshold and the percentage of incorrect predictions above this threshold reported.}
    \label{fig:APT_mom_conf}
\end{figure}

\end{appendices}
\clearpage

\bibliography{sn-bibliography}

\end{document}